%% file: main_arxiv.tex
\documentclass[10pt,twocolumn,letterpaper]{article}

\usepackage[pagenumbers]{cvpr}
\usepackage{graphicx}
\usepackage{amsmath}
\usepackage{amssymb}
\usepackage{booktabs}
\usepackage{xcolor}
\usepackage{url}
\usepackage{microtype}
\usepackage{graphicx}
\usepackage{booktabs}
\usepackage[utf8]{inputenc}
\usepackage[T1]{fontenc}
\usepackage{amsfonts}
\usepackage{amsthm}
\usepackage{nicefrac}
\usepackage{bm}
\usepackage{bbm}
\usepackage{mathtools}
\usepackage{enumitem}
\usepackage{wrapfig}
\usepackage{multirow}
\usepackage{algorithm}
\usepackage{algorithmic}
\usepackage{arydshln}
\usepackage{caption}
\usepackage{subcaption}
\usepackage{siunitx}
\usepackage[symbol]{footmisc}
\usepackage{custom_macros}
\usepackage[pagebackref,breaklinks,colorlinks]{hyperref}
\usepackage[capitalize]{cleveref}
\crefname{section}{Sec.}{Secs.}
\Crefname{section}{Section}{Sections}
\Crefname{table}{Table}{Tables}
\crefname{table}{Tab.}{Tabs.}

\def\cvprPaperID{}
\def\confName{CVPR}
\def\confYear{2022}

\begin{document}
\renewcommand*{\thefootnote}{\fnsymbol{footnote}}

\title{Deblurring via Stochastic Refinement}

\author{Jay Whang$^\dag$\footnotemark[1]
\qquad
Mauricio Delbracio$^\ddag$
\qquad
Hossein Talebi$^\ddag$
\qquad
Chitwan Saharia$^\ddag$\\

Alexandros G. Dimakis$^\dag$
\qquad
Peyman Milanfar$^\ddag$ \\[0.3em]

$^\dag$University of Texas at Austin \qquad \qquad $^\ddag$Google Research
}

\maketitle

\footnotetext[1]{This work was done during an internship at Google Research.}
\renewcommand*{\thefootnote}{\arabic{footnote}}
\setcounter{footnote}{0}

\begin{abstract}
Image deblurring is an ill-posed problem with multiple plausible solutions for a given input image. However, most existing methods produce a deterministic estimate of the clean image and are trained to minimize pixel-level distortion. These metrics are known to be poorly correlated with human perception, and often lead to unrealistic reconstructions.
We present an alternative framework for blind deblurring based on conditional diffusion models. Unlike existing techniques, we train a stochastic sampler that refines the output of a deterministic predictor and is capable of producing a diverse set of plausible reconstructions for a given input. This leads to a significant improvement in perceptual quality over existing state-of-the-art methods across multiple standard benchmarks. Our predict-and-refine approach also enables much more efficient sampling compared to typical diffusion models. Combined with a carefully tuned network architecture and inference procedure, our method is competitive in terms of distortion metrics such as PSNR. These results show clear benefits of our diffusion-based method for deblurring and challenge the widely used strategy of producing a single, deterministic reconstruction.

\end{abstract}

\section{Introduction}
\label{sec:intro}

\input{figures/fig_teaser_combined}

Image deblurring is a long-standing problem in computer vision. Various conditions such as moving objects, camera shakes, or an out-of-focus lens may contribute to blurring artifacts.
Single image deblurring is a highly ill-posed inverse problem where multiple plausible sharp images could lead to the very same blurry observation. Nonetheless, most existing methods produce a single deterministic estimate of the clean image.

Traditional methods formulate deblurring as a variational optimization problem and find a solution that satisfies closeness to certain image and/or blur kernel prior~\cite{chan1998total,fergus2006removing,shan2008high,levin2009understanding,jin2018normalized}. With the emergence of deep learning, convolutional neural networks (CNNs) have become the de-facto standard for deblurring models~\cite{tao2018scale,cho2021rethinking,tsai2021banet,zhang2019deep,li2021perceptual,kupyn2019deblurgan,suin2020spatially,shen2019human}. Typically, these CNNs are trained with simulated sharp-blurry image pairs through supervised learning. Minimizing $L_1$ or $L_2$ pixel loss is perhaps the most widely adopted approach for training such models. These losses provide a straightforward learning objective and optimize for the popular PSNR (peak signal-to-noise-ratio) metric.
Unfortunately, PSNR and other distortion metrics are well-known to only partially correspond to human perception~\cite{blau2018perception,delbracio2021projected,freirich2021theory} and can actually lead to algorithms with visibly lower quality in the reconstructed images.  To alleviate this problem, recent works introduced additional loss terms~ \cite{gatys2016image,mechrez2018contextual,mechrez2018maintaining,delbracio2021projected,kupyn2018deblurgan} that seek to improve the quality of generated images under metrics that represent human perception more reliably.
Training networks to go from corrupted images to a known ground truth in a supervised way belongs in the family of end-to-end methods~\cite{ongie2020deep}. These methods perform very well in-distribution, but can be quite fragile to distributional shifts or changes in the corruption process~\cite{ongie2020deep,jalal2021mri}.

A second body of work has focused on using deep generative models to solve inverse problems~\cite{bora2017compressed}. For deblurring, Generative Adversarial Networks (GANs)~\cite{goodfellow2014generative} have been successfully applied with competitive performance~\cite{kupyn2018deblurgan,kupyn2019deblurgan,asim2020blind}. GAN-based restoration methods train the deblurring network with an adversarial loss to make the restored images more perceptually plausible. However the proposed methods so far have been deterministic, and adversarial losses often introduce artifacts not present in the original clean image, leading to large distortion (\eg~\cite{lugmayr2021ntire} for super-resolution).

In this work, we adopt a different perspective and view deblurring as a conditional generative modeling task, where we seek to generate diverse samples from the posterior distribution.
Specifically, we introduce a \textbf{``predict-and-refine'' conditional diffusion model}, where a deterministic data-adaptive predictor is jointly trained with a stochastic sampler that refines the output of the said predictor (see \cref{fig:residual_diagram}).

Our predict-and-refine approach enables more efficient sampling compared to the standard diffusion model.
This formulation also naturally leads to a stochastic model capable of producing realistic images without sacrificing pixel-level distortion.
To the best of our knowledge, this is the first blind deblurring technique that leverages a deep generative model and is capable of producing \textit{diverse} samples.

Overall, our method produces a variety of plausible and photo-realistic results, while achieving state-of-the-art performance under many quantitative metrics in terms of both distortion and perceptual quality across multiple standard datasets.
In addition, by aggregating a different number of generated deblurred samples, our framework allows us to conveniently traverse the Perception-Distortion curve~\cite{blau2018perception,freirich2021theory} as shown in \cref{fig:teaser_combined}, without any expensive retraining or finetuning.
These results show clear benefits of stochastic diffusion-based methods for deblurring and challenge the currently dominant strategy of producing deterministic reconstructions.

\section{Related Work}
\label{sec:related_work}

\input{figures/fig_residual_diagram}

The goal of image deblurring is to generate a plausible reconstruction of the unobserved sharp, clean image $\bx$ from a blurry input $\by$.
Deblurring techniques differ in what they aim to obtain. For example, one could try to directly sample from the posterior $p(\bx \given \by)$. Another viable option is to compute a point-estimate such as the conditional mean $\E \brac{\bx \given \by}$ or the \textit{maximum a posteriori} estimate $\argmax_{\bx} p(\bx\given\by)$.

\noindent \textbf{Deblurring through point estimates}.
Traditional deblurring methods formulate the problem as one of blind deconvolution~\cite{chan1998total,fergus2006removing,shan2008high,levin2009understanding,zhu2012deconvolving,xu2013unnatural,lai2016comparative,jin2018normalized,delbracio2021polyblur,chen2019blind}. In this setup, the blur is generally modeled as a noisy linear operator acting on the clean image. While the exact values of the blur operator are not assumed to be known, one can enforce some prior distribution on the blur and the sharp image and try to find the most likely solution.

Alternatively, many recent methods adopt an end-to-end approach where a deep neural network is trained to directly produce a point estimate~\cite{chakrabarti2016neural,gao2019dynamic,nah2017deep,ramakrishnan2017deep, kupyn2018deblurgan,kupyn2019deblurgan,su2017deep,sun2015learning,tao2018scale,wieschollek2017learning, ren2021deblurring,chen2021hinet,cho2021rethinking}. These methods generally rely on pairs of blurry-sharp images as training data and cast the deblurring problem as a supervised regression task.  Much of the efforts have gone into developing specialized network architectures and loss functions to achieve better pixel-level reconstruction metrics such as PSNR or SSIM~\cite{wang2004image}.
For example, MIMO-UNet \cite{cho2021rethinking} proposed an architecture that facilitates information flow across different image resolutions in a multi-scale U-Net \cite{ronneberger2015u}. Another work HINet \cite{chen2021hinet} introduced Half Instance Normalization \cite{ulyanov2016instance}, which can be used as a building block for image restoration networks. MPRNet \cite{zamir2021multi} presented an improved multi-stage architecture designed to incorporate both high-level global features as well as local details.

\noindent \textbf{Issue of \emph{regression to the mean}}.
While the aforementioned approaches lead to state-of-the-art PSNR, they share the limitation that they can only produce a deterministic output.  This is at odds with the nature of blind image deblurring, which is an inherently ill-posed inverse problem with \textit{multiple} valid solutions for a single input.
In fact, the current trend of developing point-estimators that directly minimize a distortion loss suffers from the problem of ``regression to the mean''. If there are multiple possible clean images that correspond to the blurry input, the optimal reconstruction according to the given loss function will be an \textbf{average} of them.
Consequently, the resultant deterministic reconstruction often lacks details as it learns to produce the average of all possible solutions at best.

\noindent \textbf{Diverse image restoration}. One way to circumvent the \emph{regression to the mean} phenomenon is to avoid point estimations and directly learn to generate samples from the posterior distribution~\cite{kawar2021stochastic,kawar2021snips,ohayon2021high,kadkhodaie2021stochastic}.
While techniques based on adversarial training have been explored for blind deblurring~\cite{kupyn2018deblurgan,kupyn2019deblurgan}, in general they are not trained to produce multiple samples.
Additionally, non-reference based adversarial losses can introduce significant hallucinations and distortions~\cite{cohen2018distribution}.

Likelihood-based deep generative models such as
Variational Autoencoders~\cite{prakash2020fully}, Normalizing Flows~\cite{lugmayr2020srflow,lugmayr2021ntire}, and Diffusion Probabilistic Models (DPMs)~\cite{saharia2021image,li2021srdiff} have also been successfully applied to other image enhancement tasks such as super-resolution, where a diverse set of candidates can be generated from the learned posterior~\cite{prakash2020fully}.
Compared to point estimates, solving imaging inverse problems by sampling from the posterior has additional benefits such as uncertainty quantification~\cite{kawar2021stochastic,whang2021composing,kawar2021snips}, near-optimal sample complexity~\cite{jalal2020robust} and better fairness guarantees~\cite{jalal2021fairness}.

\section{Diffusion Probabilistic Models}
\label{sec:background}

Diffusion probabilistic model \cite{sohl-dickstein2015deep,ho2020denoising} is a latent variable model specified by a $T$-step Markov chain $(\bx_0, \bx_1, \ldots, \bx_T)$ called the \textit{diffusion process}. It starts from a clean data sample $\bx_0 \in \R^d$ and repeatedly injects Gaussian noise according to the transition kernel $q(\bx_{t} \given \bx_{t-1})$ as follows:
\begin{align}
    q(\bx_{t} \given \bx_{t-1}) &\triangleq \N(\bx_{t}; \sqrt{\alpha_t}\bx_{t-1}, (1-\alpha_t)\I),
\end{align}
where $\alpha_t \in (0,1)$ for all $t=1,\ldots,T$. The noise schedule $\balpha_{1:T} \triangleq (\alpha_1, \ldots, \alpha_T)$ is a hyperparameter that controls the variance of noise added at each step. The latent variables $\bx_{1:T}$ have the same dimensionality as the original data sample $\bx_0$.

While this particular choice of diffusion process may seem arbitrary, it results in closed-form expressions for the following distributions: the marginal\footnote{For notational brevity, we use the term ``marginal'' to include distributions conditioned on $\bx_0$.} distribution $q(\bx_t \given \bx_0)$ and the \textit{reverse diffusion step} $q(\bx_{t-1} \given \bx_t, \bx_0)$. Writing $\baralpha_t \triangleq \prod_{j=1}^t \alpha_j$, we get
\begin{align}
\label{eqn:forward_marginal}
    q(\bx_t \given \bx_0) &= \N(\bx_t; \sqrt{\bar{\alpha}_t}\bx_0, (1-\bar{\alpha}_t)\I) \\
\label{eqn:forward_posterior}
    q(\bx_{t-1} \given \bx_t, \bx_0) &= \N(\bx_{t-1}; \bmu_t(\bx_t,\bx_0), \beta_t\I),
\end{align}
where $\bmu_t(\bx_t,\bx_0)$ and $\beta_t$ are quantities that depend on $\bx_t, \bx_0$ and $\balpha_{1:T}$. Their full expressions and derivations are included in \Cref{supp:dpm}.

The marginal distribution in \cref{eqn:forward_marginal} allows us to sample a partially noisy image $\bx_t$ at an arbitrary time step, and the reverse diffusion step in \cref{eqn:forward_posterior} is a stochastic denoising procedure that tells us how to reverse a single diffusion step by sampling a slightly less noisy image $\bx_{t-1}$ from $\bx_t$. The ability to sample from arbitrary marginals is important to make training of a DPM practical, as the training objective relies on it (see \cref{eqn:base_loss}).

We note that the diffusion process defined here has no learnable parameter.  It is a fixed process that gradually destroys the original signal $\bx_0$ and produces $\bx_T$ that looks indistinguishable from pure Gaussian noise given a sufficiently large $T$.
Thus, if we could apply the reverse diffusion step $T$ times starting from pure Gaussian noise, we would obtain a clean sample $\bx_0$. However this is not possible because the reverse diffusion step itself requires access to $\bx_0$, which is exactly what we are trying to generate.

\noindent \textbf{Reverse process and denoiser network}.
A key component of DPM is the \textit{denoiser network} $\ftheta$ that tries to estimate $\bx_0$ from the partially noisy image $\bx_t$. With it, we can apply the reverse diffusion step without knowing $\bx_0$ by using the estimate $\ftheta(\bx_t, t)$ in place of $\bx_0$:
\begin{equation}
    \label{eqn:reverse_process}
    \ptheta(\bx_{t-1}\given\bx_t) \triangleq q(\bx_{t-1} \given \bx_t, \ftheta(\bx_t, t))
\end{equation}

This defines a Markov chain that runs backwards in time from $\bx_T$ to $\bx_0$, which we call the \textit{reverse process}. The goal of DPM is to train $\ftheta$ to make $\ptheta(\bx_{t-1}\given\bx_t)$ as close to the true reverse diffusion step $q(\bx_{t-1} \given \bx_t, \bx_0)$ as possible. This is done by optimizing $\ftheta$ to maximize the variational lower bound of the marginal likelihood $\log \ptheta(\bx)$.

In practice, we use an alternative parametrization of $\ftheta$ proposed by~\cite{ho2020denoising} that instead predicts the Gaussian noise $\eps$ that deterministically relates $\bx_t$ and $\bx_0$ via \Cref{eqn:forward_marginal}. Specifically, we write $\bx_t = \sqrt{\baralpha_t} \bx_0 + (1-\baralpha_t) \eps$ for $\eps \sim \N(\bzero, \I)$ and train $\ftheta$ to predict $\eps$.

\noindent \textbf{Continuous noise level}.
Chen \etal \cite{chen2020wavegrad} proposes a modified formulation based on a continuous noise level $\baralpha$, which we also adopt.
An important property of this formulation is that it allows us to sample from the model using a noise schedule $\balpha_{1:T}$ different from the one used during training.  This flexibility enables us to control the trade-off between the distortion and the perceptual quality of generated samples without having to retrain the model, as we show later.

\noindent \textbf{Conditional DPM}.
So far we have defined a DPM that is trained to model the \textit{unconditional} data distribution.
For conditional models that must estimate $p(\bx\given\by)$, we make $\ftheta$ accept $\by$ as the conditioning input, as was done in~\cite{saharia2021image,chen2021wavegrad}. This way, the iterative denoising procedure becomes dependent on $\by$. The final training objective is:
\begin{equation}
    \label{eqn:base_loss}
    L_{\text{Base}}(\theta) = \E \norm{\eps - \ftheta(\sqrt{\baralpha} \bx_0 + \sqrt{1-\baralpha}\eps, \baralpha, \by)}_1,
\end{equation}
where the expectation is over $\by, \bx_0, \baralpha$, and $\eps$.

\noindent \textbf{Sampling from a DPM}.
As mentioned earlier, sampling an image from a DPM is done by running the reverse process.  Given some inference-time noise schedule $\baralpha_{1:T}$, we start from a pure Gaussian noise $\bx_T \sim \N(\bzero, \I)$ and repeatedly apply the reverse process transition $\ptheta(\bx_{t-1}\given\bx_t)$ defined in \cref{eqn:reverse_process}.  Notice that this procedure requires a total of $T$ calls to the denoiser network. At the end of this sampling procedure, we are left with a single sample $\bx_0$.

\section{Predict-and-Refine Diffusion Model}
\label{sec:our_approach}

\input{figures/fig_proposal_and_samples}

One of the main drawbacks of DPM is the computational cost of generating samples, which may require up to thousands of forward passes of the denoiser network due to the iterative denoising procedure. As such, many recent works have explored alternative sampling strategies that reduce the number of sampling steps \cite{song2021denoising,san2021noise,jolicoeur2021gotta,kong2021on,watson2021learning,lee2021priorgrad}.

We introduce a simple technique that reduces this cost by exploiting the fact that it is often possible to get a cheap initial guess for conditional generative models.  Specifically, we augment our conditional diffusion model with a deterministic \mbox{\textit{initial predictor}} (\cref{fig:residual_diagram}), which provides a data-adaptive candidate for the clean image. Then the denoiser network only needs to model the \textit{residual}.

Letting $\gtheta$ denote the initial predictor, the new objective becomes:
$L_\text{Ours}(\theta) =$
\begin{align}
    \label{eqn:our_loss}
    &\E \norm{\eps - \ftheta\Big(\sqrt{\baralpha} \big(\underbrace{\bx_0 - \gtheta(\bx_0)}_{\text{residual}}\big) + \sqrt{1-\baralpha}\eps,\baralpha,\by\Big)}_1
\end{align}
We include a pseudocode for the modified sampling procedure in \cref{alg:residual_sampling}. Notice that the initial predictor $\gtheta$ does not require an extra loss or pretraining because the gradient from the loss flows through $\ftheta$ into $\gtheta$.

Since the initial predictor runs only once, it is beneficial to keep the denoiser network small by offloading most of the computation to the initial predictor.
This leads to much more efficient sampling because any reduction in the computational cost of the denoiser network gets amplified by the number of sampling steps used. We further explore this effect in \cref{sec:analysis}.

\input{algorithms/residual_sampling}

\subsection{Perception-Distortion Trade-off}

As explained in \cref{sec:background}, conditioning the diffusion model on continuous noise level makes it possible to use a different noise schedule during inference.
We observe that using many steps with small noise level generally leads to better perceptual quality, and using fewer steps with large noise level leads to lower distortion.

For our experiments, we run a small grid search over the noise schedule hyperparameters and use the model with the best LPIPS score (labeled ``Ours'').  We emphasize that this inference-time hyperparameter tuning is cheap as it does not involve retraining or finetuning the model itself.

\noindent \textbf{Sample averaging}.
Our framework also provides a principled alternative to geometric self-ensemble \cite{lim2017enhanced}. Since our stochastic sampler is trained to learn the target posterior $p(\bx\given\by)$, we can average multiple samples from our model to approximate the conditional mean $\E\brac{\bx \given \by}$, \ie the minimum mean squared error estimator. We thus report results for a second model (labeled ``Ours-SA'') that returns the average of multiple samples.

\noindent \textbf{Traversing the Perception-Distortion curve}.
By appropriately setting the inference-time hyperparameters mentioned above (sampling steps $T$, noise schedule $\baralpha_{1:T}$, and sample averaging), we can smoothly traverse the P-D curve as shown in \cref{fig:teaser_combined}.

For example, the LPIPS-optimized model (``Ours'') uses a relatively large step count of $T=500$ \textit{without} sample averaging to achieve high perceptual quality at a slight cost of PSNR. The distortion-optimized model (``Ours-SA'') does the opposite by using $T=10$ with sample averaging to sacrifice perceptual quality for higher PSNR. Each point on the P-D curve in \cref{fig:teaser_combined} thus corresponds to a specific choice of these hyperparameters.

\subsection{Resolution-agnostic Architecture}
Unlike the image benchmarks commonly used to evaluate DPMs, blind deblurring benchmarks contain images with various sizes. To support arbitrary input shapes, we use a fully-convolutional architecture for both initial predictor and denoiser network.

Our architecture is based on SR3 \cite{saharia2021image}, which uses a variant of U-Net architecture from \cite{ho2020denoising} with residual blocks replaced with that of BigGAN \cite{brock2018large}.  To make our model agnostic to image resolution, we removed self-attention, positional encoding, and group normalization. The exact specification of our architecture can be found in \Cref{supp:model_details}.

We note that, to the best of our knowledge, this is the first time a conditional diffusion model is made to support arbitrary image size.
Our preliminary experiments show that the fully-convolutioanl architecture had little to no degradation in sample quality for deblurring at non-native resolutions.
Because the denoiser network is a relatively simple U-Net, DPMs provide a particularly convenient choice for conditional image generation that must work on any input size.

\section{Experiments}
\label{sec:experiments}

\subsection{Datasets}

We train and evaluate our models on two widely-used image deblurring datasets. For a fair comparison, we follow the same setup used by \cite{nah2017deep,kupyn2019deblurgan,chen2021hinet,cho2021rethinking,suin2020spatially,zhang2019deep} and train our model only using the provided training data.

\noindent \textbf{GoPro}.
GoPro dataset \cite{nah2017deep} contains 3214 pairs of clean and blurry $1280 \times 720$ images, of which 1111 are reserved for evaluation. These images are generated by recording video clips with high shutter speed, then averaging consecutive frames to simulate blurs caused by slow shutter speed.

\noindent \textbf{HIDE}.
We additionally evaluate our GoPro-trained model on the HIDE \cite{shen2019human} dataset, which contains 2025 images also of size $1280 \times 720$. By training and evaluating our model on different datasets, we can test its ability to generalize under a distributional shift.

\subsection{Model Training}

We jointly train the initial predictor and denoiser network by minimizing the loss in \cref{eqn:our_loss}.  Since our model is fully convolutional, we use random $128 \times 128$ crops during training, but apply the model on full-size images for evaluation.
We also perform training-time data augmentation with random horizontal/vertical flips and \ang{90}/\ang{180}/\ang{270} rotations.

\noindent \textbf{A note on training data}.
Most currently leading methods only report distortion-based metrics (PSNR and SSIM) and provide pre-trained models for GoPro.  Since our work focuses on perceptual quality, we need to compute perceptual metrics ourselves using outputs from other methods. Thus to ensure a fair comparison, we are limited to using models trained on the GoPro dataset, as it is the only dataset with widely available pre-trained models.  Nonetheless, we provide additional results and the details of how we obtained the outputs of other methods in \Cref{supp:additional_results,supp:evaluation_details}.

\input{tables/table_gopro_results_concise}

\subsection{Evaluation}

\noindent \textbf{Evaluation Metrics}.
We evaluate our method on four different perceptual metrics: LPIPS \cite{zhang2018perceptual}, NIQE \cite{mittal2012making}, FID (Fréchet Inception Distance) \cite{heusel2017gans}, and KID (Kernel Inception Distance) \cite{binkowski2018demystifying}. Because our datasets do not have enough examples to reliably compute FID and KID, we extract 15 non-overlapping patches of size $256 \times 240$ from each $1280 \times 720$ image and compute the Inception-based metrics at the patch level, similar to \cite{mentzer2020high}.
For completeness, we also include two distortion-based metrics: PSNR and SSIM \cite{wang2004image}.

We note the importance of including full-reference metrics for conditional image generation.  A method can achieve near-perfect score on a no-reference metric such as NIQE by producing highly realistic images that are completely unrelated to the input. This is particularly relevant for GAN-based methods, since the discriminator may not penalize the generator for producing natural-looking images that do not match the input.  This is why we included LPIPS (and to some extent, PSNR and SSIM), even though it is technically not a perceptual metric.
For a qualitative comparison, we also conduct a human study and provide sample restorations.

\input{figures/fig_gopro_and_hide_samples_arxiv}

\subsection{Quantitative Results}

\subsubsection{GoPro Results}
\Cref{table:gopro_results} shows quantitative results on the GoPro dataset. We compared our model with the current state-of-the-art (SOTA) methods HINet~\cite{chen2021hinet}, MPRNet~\cite{zamir2021multi}, and DeblurGAN-v2~\cite{kupyn2019deblurgan}.

Our model achieves SOTA performance across \textbf{all perceptual metrics} while maintaining competitive PSNR and SSIM to existing methods. Notably, we obtain the FID of 4.04, \textbf{nearly a 70\% reduction} compared to DeblurGAN-v2 \cite{kupyn2019deblurgan}, the current SOTA method in terms of perceptual quality.
Moreover, the sample-averaging variant of our method achieves a new SOTA PSNR of 33.23 while still outperforming all other methods with respect to LPIPS.
All in all, these results highlight our framework's flexibility to control the trade-off between perception and distortion using a single model.
As shown in \Cref{fig:teaser_combined}, our result sets a new Pareto frontier on the Perception-Distortion plot.

\input{tables/table_hide_results_concise}

\subsubsection{HIDE Results}
We also evaluate our GoPro-trained model on the HIDE dataset \cite{shen2019human} to test its ability to generalize to out-of-distribution input. As the results in \Cref{table:hide_results} clearly show, the gains in perceptual quality do translate over to the HIDE dataset.  In particular, both of our models significantly outperform the baseline methods across \textbf{all perceptual metrics} while maintaining competitive distortion values.

\cref{fig:gopro_and_hide_samples} includes several sample reconstructions from both GoPro and HIDE datasets.  Despite sometimes containing a little more noise (some of which was presumably learned from the training data itself), we see that our model shows a clear improvement in perceptual quality. Additional full-size comparisons are provided in \Cref{supp:large_results}.

\subsection{Human Study for Qualitative Evaluation}
\label{sec:human_study}

We ran a perceptual study with human subjects to further quantify the performance of the proposed deblurring framework. Our results are presented in \Cref{table:perceptual_study}. We used Amazon Mechanical Turk to obtain pairwise ratings comparing different deblurring methods applied on the GoPro dataset. In this study, the human subjects had a minimum of 70\% approval rating, and  were  asked to select the image with the better quality from side-by-side crops of size $512\times512$.

Results in \Cref{table:perceptual_study} show the average rater's preference computed from 480 comparisons. As the highlighted cells show, these results indicate that \textbf{both variations of our deblurring model outperform the competing methods}.

We also observed that raters showed a modest preference for the sample-averaged variant in crops with relatively flat content. On the other hand, raters preferred individual samples for highly-textured crops.
\cref{fig:diversity_std} shows that the level of detail produced by our model is adaptive to the blur present in the input.  As expected, blurrier images generally lead to higher variance in the resulting samples.

\input{tables/table_perceptual_study}

\section{Discussion and Analysis}
\label{sec:analysis}

For the analysis of various aspects of our model, we used a custom dataset created by applying synthetic camera shake blur and noise (described in \Cref{supp:div2k_dataset})
on the images of the DIV2K dataset \cite{agustsson2017ntire}. This was done to make qualitative evaluation in a more controlled environment, since the low-quality ground truth images in existing paired datasets \cite{nah2017deep,rim2020real} make qualitative assessment difficult.

\input{figures/fig_diversity_std}

\subsection{Benefits of Residual Modeling}

\noindent \textbf{More efficient sampling.}
The main benefit of residual modeling is the reduction in the computational cost of sampling.  Due to the iterative nature of diffusion sampling, the denoiser network must run many times for each generated sample -- sometimes up to hundreds to thousands of steps.  Thus, any reduction in the cost of running the denoiser is particularly valuable,
and our initial predictor provides a simple way to offload some of this computation.

A key question is then whether the initial predictor can compensate for the decrease in the sample quality from using a smaller denoiser network.  We empirically explore this by comparing sampling latency against sample quality with and without the initial predictor.
In \cref{fig:sampling_cost}, the non-residual model refers to a regular conditional diffusion model with a large denoiser network.  The residual model follows our architecture and has a large initial predictor and a small denoiser. Overall, the residual model has more parameters (33M vs. 28M).

We see that the residual model requires much less time to sample an image despite it being larger than the non-residual model.
Importantly, this reduction in sampling cost does not negatively affect the sample quality -- in fact, the residual model is up to $7\times$ faster for a comparable sample quality.

\input{figures/fig_sampling_cost}

\noindent \textbf{Output of the initial predictor}.
One unexpected discovery from our experiments is that the output of the initial predictor is often a fairly reasonable reconstruction of the reference image. We can see this in \cref{fig:proposal_and_samples}. While lacking in detail, the initial prediction is certainly less blurry than the input.

It is perhaps surprising that this happens even though there is no explicit loss on the initial predictor's output $\gtheta(\by)$ to match the reference. We also note that our method is not the only possible parameterization of a diffusion model with an explicit decoupling of the iterative portion (denoiser network) from the single-pass portion (initial predictor).
For instance, we could have simply fed $\gtheta(\by)$ as an auxiliary input to the denoiser $\ftheta$ without computing the residual.
We leave these investigations around the initial predictor as future work.

\noindent \textbf{Residual images are simpler to model}.
One may wonder why adding a deterministic initial predictor would help with the model's performance.
We posit that the benefits of residual modeling may be due to the distribution of residual images being ``simpler'' than that of reference images.

While it is impractical to approximate the true entropy of the two distributions, we can look at related quantities that may serve as a proxy.
Specifically, we compute the entropy of pixel values aggregated across all pixel locations for residual and reference images.
As expected from natural images, the reference pixel distribution is reasonably spread out and has the entropy of $7.42$ bits-per-dimension (bpd).  On the other hand, the residual pixel values follow a much more sharply concentrated distribution, leading to a substantially lower entropy of $3.91$ bpd. This suggests that the residual images may indeed be simpler to model.

\subsection{Network Architecture Ablation}
To better understand where the performance gains of our method are originating from, we trained a regression-based baseline that only uses the initial predictor.  Surprisingly, we observed that the initial predictor alone was able to achieve state-of-the-art PSNR of 33.07 when trained with a simple $L_2$ loss.  Through a detailed ablation study, we identified three key hyperparameters: exponential moving average (EMA) of weights, large batch size, and network size.

In \Cref{table:regressor_ablation}, we start from a simple U-Net architecture \cite{ronneberger2015u} and gradually enable each of the aforementioned hyperparameters. All models were trained for 1M steps to ensure the differences are not due to insufficient training. As the results show, all three hyperparameters were critical to the model's performance.

\input{tables/table_regressor_ablation_concise}

\section{Conclusion and Future Directions}

We presented a new framework for stochastic blind image deblurring with a focus on perceptual quality using a conditional diffusion model. We introduced a novel technique for reducing the computational burden of diffusion sampling.  We empirically showed that our method achieves significantly improved perceptual quality and competitive distortion metrics as compared to the current state-of-the-art methods. We believe that our work opens a new direction for blind deblurring with a focus on perceptual quality and establishes a strong benchmark for future works to improve upon.

There are a number of avenues to explore to further address the limitations of our work. Due to slow sampling and large network size, diffusion models are computationally too expensive to be incorporated into consumer-level devices. One way to combat this is to use more efficient sampling schemes such as DDIM \cite{song2021denoising} or distillation \cite{anonymous2022progressive}. Another promising direction is to replace our initial predictor and denoiser network with U-Net architectures that are optimized for both distortion and run time \cite{cho2021rethinking,zamir2021multi,chen2021hinet}.

{\small
\bibliographystyle{ieee_fullname}
\bibliography{egbib}
}

\clearpage
\appendix
\begin{center}
{\Large \bf Appendix}
\end{center}

\section{Additional Perception-Distortion Plots}
The Perception-Distortion plot provided in Section~1 of the main text shows the trade-off between PSNR and Kernel Inception Distance (KID).  We observe that other combinations of perceptual (NIQE, LPIPS, FID) and distortion metrics (PSNR, SSIM) follow a similar trend, as shown in \Cref{fig:additional_pd_plots}. We note that formally LPIPS is also a distortion metric, as it is a full-reference based distance computed in a deep feature space. We nonetheless observed that LPIPS corresponds to human perception much better than PSNR or SSIM.

\input{figures/sm/fig_additional_pd_plots}

\section{Diversity Analysis}

\Cref{fig:diversity_std_blur_gopro} shows the relation between the blurriness (or sharpness) on the input image, and the diversity of the generated deblurred samples. The blurrier the input image is, the more diversity we get in the samples (see figure caption for more details).

\input{figures/sm/fig_diversity_std_blur_gopro}

\section{Synthetic DIV2K Deblurring Dataset}
\label{supp:div2k_dataset}
To better analyze various aspects of our diffusion deblurring model, we created a custom dataset by applying synthetic camera shake blur (following~\cite{delbracio2015removing} and noise to the DIV2K dataset~\cite{agustsson2017ntire}.
This allows us to make qualitative evaluations in a more controlled environment, since the low-quality ground truth images in existing paired datasets \cite{nah2017deep,rim2020real} make qualitative assessment difficult and lessens the benefits from using a powerful generative model.
\begin{figure}
    \centering
    \includegraphics[width=1.0\linewidth]{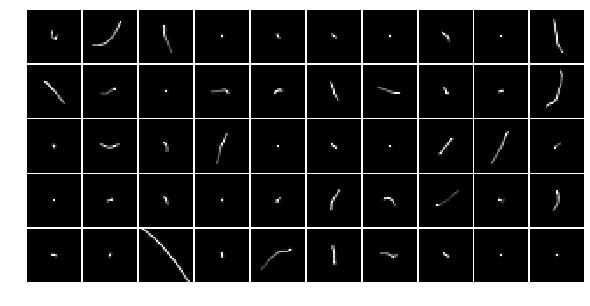}
    \caption{Examples of synthetically generated random kernels (following~\cite{delbracio2015removing}) used to generate the deblurring dataset.}
    \label{fig:kernels}
\end{figure}

The synthetically generated random kernels are of varying size ($31 \times 31$ maximal support). Figure~\ref{fig:kernels} shows example kernels. The kernels can be of any size from a perfect Delta (sharp) to about 30 pixels. In addition to the blur, a white Gaussian noise with random standard deviation $\sigma \sim \mathcal{U}[0,15]$ is added.

\newcommand\myeq{\stackrel{\mathclap{\normalfont\mbox{def}}}{=}}

\section{Omitted Details for DPM Formulation}
\label{supp:dpm}

\noindent \textbf{\Cref{eqn:forward_marginal}: Marginal at time step $t$.}
We proceed by induction. For $t=1$, we have $\baralpha_1 = \alpha_1$, so \cref{eqn:forward_marginal} reduces to the diffusion transition kernel:
\begin{equation*}
    q(\bx_{1} \given \bx_{0}) = \N\paren{\bx_{1}; \sqrt{\alpha_1}\bx_{0}, (1-\alpha_1)\I}.
\end{equation*}
Now suppose we have
$q(\bx_t \given \bx_0) = \N(\bx_t; \sqrt{\bar{\alpha}_t}\bx_0, (1-\bar{\alpha}_t)\I)$ for some $t > 1$,
which we reparameterize as
\begin{equation*}
    \bx_{t} = \sqrt{\baralpha_{t}} \bx_{0} + \sqrt{1-\baralpha_{t}} \eps, \text{ where } \eps \sim \N(\bzero, \I).
\end{equation*}
Then by applying a single diffusion step $q(\bx_{t+1}\given \bx_t)$ to the above, we get
\begin{align*}
    \bx_{t+1}
    &\stackrel{(1)}{=} \sqrt{\alpha_{t+1}} \bx_{t} + \sqrt{1-\alpha_{t+1}} \eps' \\
    &\stackrel{(2)}{=} \sqrt{\alpha_{t+1}}\sqrt{\baralpha_t} \bx_0 + \sqrt{\alpha_{t+1}}\sqrt{1-\baralpha_t}\eps + \sqrt{1-\alpha_{t+1}} \eps'\\
    &\stackrel{(3)}{=} \sqrt{\baralpha_{t+1}} \bx_0 + \sqrt{\alpha_{t+1}-\baralpha_{t+1}}\eps + \sqrt{1-\alpha_{t+1}} \eps' \\
    &\stackrel{(4)}{=} \sqrt{\baralpha_{t+1}} \bx_0 + \sqrt{1-\baralpha_{t+1}} \eps'',
\end{align*}
where the first step uses a reparameterization $\eps' \sim \N(\bzero, \I)$, the second step is from the inductive hypothesis, and the  last step follows from summing two independent Gaussian random variables.
Thus
\begin{equation*}
    \bx_{t+1} \sim \N\paren{\sqrt{\baralpha_{t+1}}\bx_0, (1-\baralpha_{t+1})\I},
\end{equation*}
which concludes the inductive step.

\vspace{1em}
\noindent \textbf{Reverse diffusion step expressions.}
Applying Bayes' Rule to \cref{eqn:forward_posterior} leads to the following expressions for the mean and variance for the reverse diffusion step:
\begin{align*}
    &\bmu_t(\bx_t, \bx_0) = \frac{\sqrt{\baralpha_{t-1}}(1-\alpha_t)}{1-\baralpha_t} \bx_0 +
    \frac{\sqrt{\alpha_t}(1-\baralpha_{t-1})}{1-\baralpha_t} \bx_t, \\
    &\beta_t = \frac{1-\baralpha_{t-1}}{1-\baralpha_{t}}(1-\alpha_t).
\end{align*}
We refer the reader to Ho \etal~\cite{ho2020denoising} for a more thorough treatment of the DPM formulation.

\vspace{1em}
\noindent \textbf{Specifying the noise schedule.}
Following \cite{chen2020wavegrad,saharia2021image}, given a fixed budget of $T$ steps, we sample the \textit{continuous} noise level $\sqrt{\baralpha}$ from a piecewise uniform distribution.
Specifically, we define $T$ intervals $(l_{i-1}, l_{i})$, where $l_0 \triangleq 1$ and $l_i \triangleq \sqrt{\baralpha_i}$ for $i > 0$.  Then to sample a continuous noise level $\baralpha$, we first randomly pick an interval $(l_{k-1}, l_k)$, and sample $\baralpha \sim \mathcal{U}[l_{k-1}, l_k]$.

Now all that remains is to specify the schedule $\alpha_1, \ldots, \alpha_T$.  While there are many options (\eg as explored by Chen \etal \cite{chen2020wavegrad}), we used a simple linear schedule on the variance of the forward process by fixing the two endpoints and linearly interpolating the intermediate values.

\section{Model Details}
\label{supp:model_details}

\noindent \textbf{Network architecture.}
We use a U-Net \cite{ronneberger2015u} architecture similar to the one used by SR3 \cite{saharia2021image}.  A crucial difference is that our network was made fully-convolutional by removing self-attention, group normalization, and positional encoding.  At the input, the noisy sample $\bx_t$ is concatenated with the conditioning input $\by$ channel-wise.

As shown in \cref{fig:architecture_diagram}, our U-Net has four resolution depths with channel multipliers $\set{1, 2, 3, 4}$.  Both the denoiser network and initial predictor use this architecture. Their main difference is size, where the starting channel count is 64 for the initial predictor and 32 for the denoiser.  This results in the initial predictor having $\sim$26M parameters, and the denoiser having $\sim$7M parameters.
Note that the input and output would change slightly when this architecture is used for the initial predictor, which tries to estimate $\bx$ from $\by$ (no $\bx_t$ and $\baralpha$ in the input, and the output is not $\eps$).

\vspace{1em}
\noindent \textbf{Training details.}
We train all of our models for 1M steps using 32 TPUv3 cores.  For our main model with the initial predictor and the denoiser network, it takes about 27 hours to train the model.  We used the AdamW \cite{loshchilov2018decoupled} optimizer with a fixed learning rate of 0.0001, weight decay rate of 0.0001, and EMA decay rate of 0.9999.
During training, we used fine-grained diffusion process with $T=2000$ steps.  As described above, we used a linear noise schedule with the two endpoints set as:
 $1-\alpha_0 = 1\times 10^{-6}$ and $1-\alpha_T = 0.01$.

\input{figures/sm/fig_architecture_diagram}

\section{Evaluation Details}
\label{supp:evaluation_details}

For all our experiments (on all datasets: GoPro, HIDE, DIV2K), we performed a grid search over the following hyperparameter combinations during inference:
\begin{enumerate}
    \item Inference steps ($T$): 10, 20, 30, 50, 100, 200, 300, 500.
    \item Noise schedule ($\balpha_{1:T}$): We fixed the initial forward process variance ($1 - \alpha_0$) to $1\times10^{-6}$.  For the final variance ($1-\alpha_T$), we sweep over $\set{0.01, 0.02, 0.05, 0.1, 0.2, 0.5}$.  The intermediate values are linearly interpolated.
\end{enumerate}

\noindent \textbf{How baseline samples are obtained.}
As mentioned in \cref{sec:experiments} of the main text, we computed various perceptual metrics ourselves as the existing literature often only reports PSNR and SSIM.  To ensure fairness in our comparisons, we tried to use author-produced restoration results whenever possible.  Otherwise, we used the official implementations and pre-trained models released by the authors of each paper and produced restorations ourselves.

Specifically, for HINet \cite{chen2021hinet}, MPRNet \cite{zamir2021multi}, and SAPHNet \cite{suin2020spatially}, we used restorations produced by the authors for both GoPro and HIDE results.
For MIMO-UNet+ \cite{cho2021rethinking} and DeblurGANv2 \cite{kupyn2019deblurgan}, we used the authors' implementation and model checkpoints from their respective Github repositories.
For SimpleNet \cite{li2021perceptual}, we could not obtain either the restorations nor the code, so we only reported the metrics from the paper (PSNR, SSIM, LPIPS).

\section{Large GoPro and HIDE Results}
\label{supp:large_results}

In Figures \ref{fig:supp_gopro_fullsize_1}--\ref{fig:supp_hide_fullsize_1}, we include larger versions of the GoPro and HIDE restorations shown in the main text.  Figures \ref{fig:supp_gopro_fullsize_1} and \ref{fig:supp_gopro_fullsize_2} are from GoPro~\cite{nah2017deep}, and Figure \ref{fig:supp_hide_fullsize_1} is from HIDE dataset \cite{shen2019human}.

\input{figures/sm/fig_gopro_fullsize_1}
\input{figures/sm/fig_gopro_fullsize_2}
\input{figures/sm/fig_hide_fullsize_1}

\section{Additional Results}
\label{supp:additional_results}

\noindent \textbf{GoPro dataset.} In Figures \ref{fig:supp_gopro_samples_1}--\ref{fig:supp_gopro_samples_5} we present additional results on the GoPro dataset~\cite{nah2017deep} where we compare our diffusion deblurring method to SAPHNet~\cite{suin2020spatially}, DeblurGAN-v2~\cite{kupyn2019deblurgan}, MIMO-Unet+~\cite{cho2021rethinking}, MPRNet~\cite{zamir2021multi}, and HINet~\cite{chen2021hinet}.  Consistent with the main text, ``Ours-SA'' refers to the sample averaging variant of our method.

\input{figures/sm/fig_gopro_samples_1}
\input{figures/sm/fig_gopro_samples_2}
\input{figures/sm/fig_gopro_samples_3}
\input{figures/sm/fig_gopro_samples_4}
\input{figures/sm/fig_gopro_samples_5}

\noindent \textbf{DIV2K Deblurring dataset.} In Figures \ref{fig:supp_div2k_samples_1}--\ref{fig:supp_div2k_samples_4} we present additional results on the synthetically generated DIV2K deblurring dataset. For comparison purposes, we train a regression-based model (to minimize L2 loss, thus maximizing PSNR) that has the same architecture as the one we used for the initial predictor. Compared to the over-smoothed restorations from the regression-based baseline trained to minimize distortion, our method produces more realistic textural details.

\input{figures/sm/fig_div2k_samples_1}
\input{figures/sm/fig_div2k_samples_2}
\input{figures/sm/fig_div2k_samples_3}
\input{figures/sm/fig_div2k_samples_4}

\end{document}

%% file: figures/fig_teaser_combined.tex
\begin{figure}
    \centering
    
    \begin{subfigure}[t]{\linewidth}
        \centering
        \includegraphics[width=\linewidth]{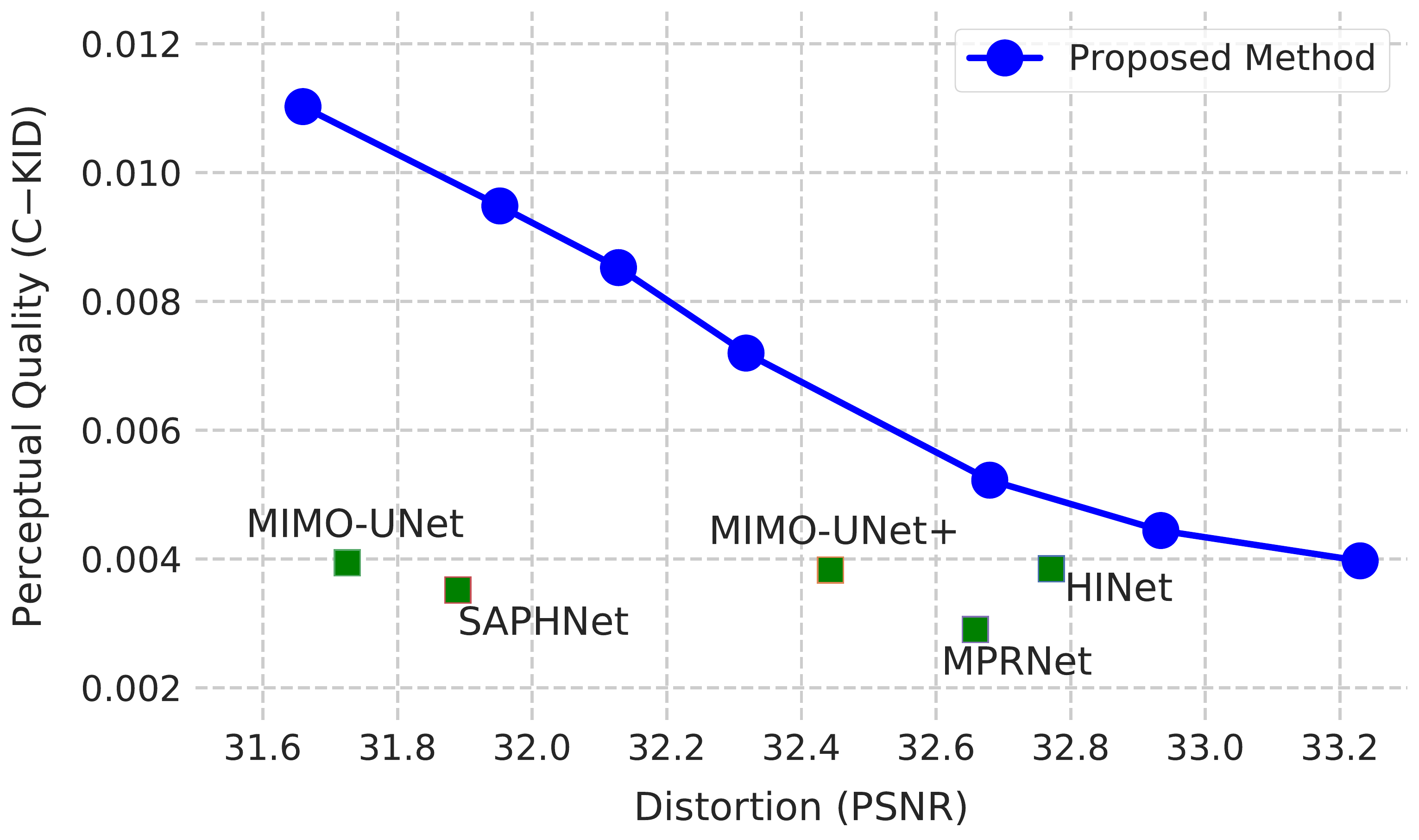}
        \label{fig:pd_curve}
    \end{subfigure}
    
    \vspace{-0.5em}
    \begin{subfigure}[t]{\linewidth}
        \centering
        \includegraphics[width=\linewidth]{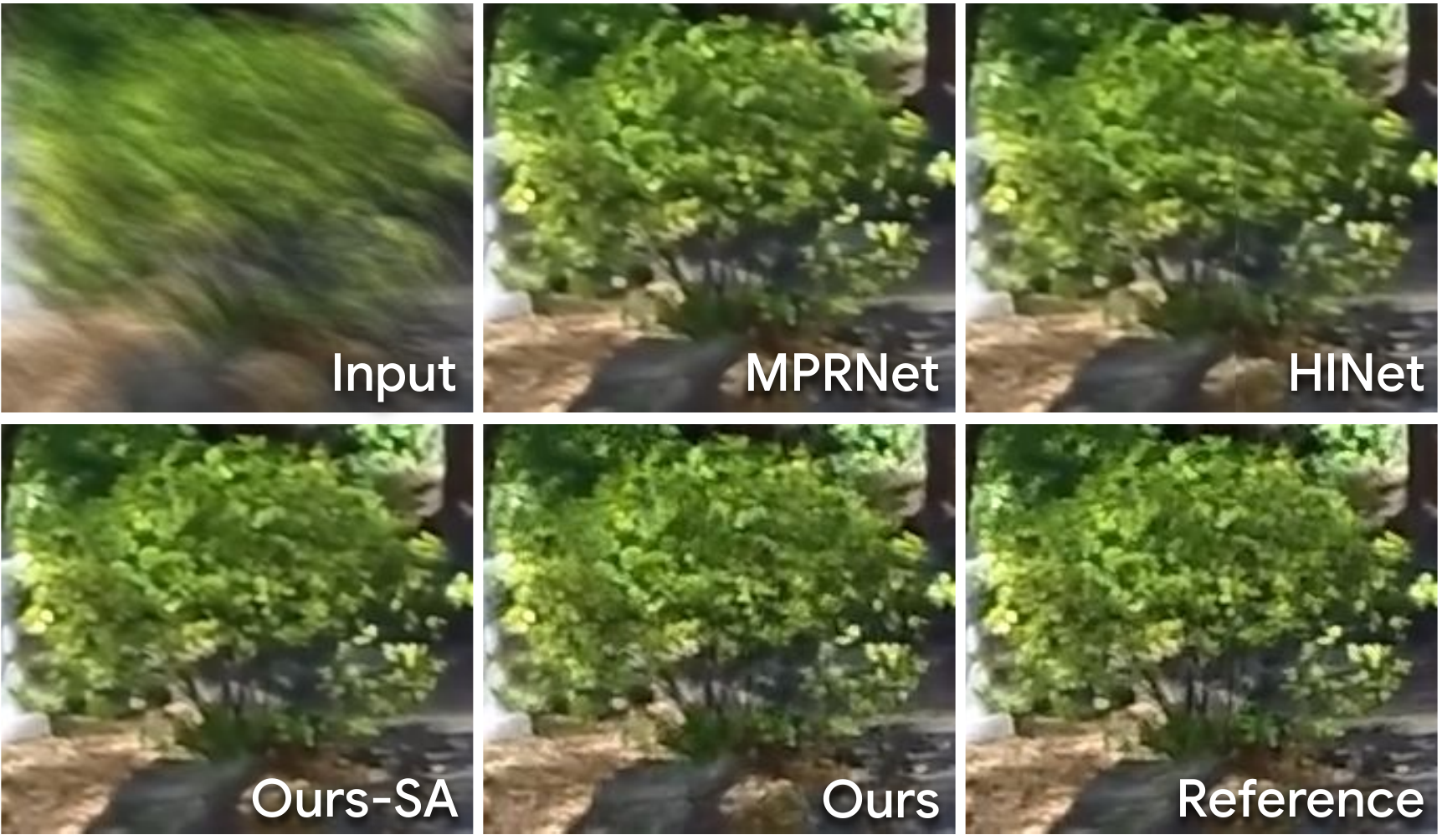}
        \label{fig:teaser}
    \end{subfigure}
    
    \vspace{-1em}
    \caption{\textbf{Top:} Perception-Distortion (P-D) trade-off \cite{blau2018perception} of current state-of-the-art deblurring methods (top).  
    Our method sets a new Pareto frontier in the P-D plot and allows us to traverse through the P-D curve using a single model without retraining or finetuning.
    \textbf{Bottom:} Samples from our method compared to other competitive methods. We include two extremes from our model -- one optimized for perceptual quality (``Ours'') and one for distortion using Sample Averaging (``Ours-SA''). These correspond to the two end points of the P-D curve. 
    For the ease of interpretation, we used negative Kernel Inception Distance \cite{binkowski2018demystifying} ($C - \text{KID}$ for a constant $C$) as the measure of perceptual quality.
    }
    \label{fig:teaser_combined}
    \vspace{-1em}
\end{figure}

%% file: figures/fig_residual_diagram.tex
\begin{figure}
    \centering
    \includegraphics[width=\linewidth]{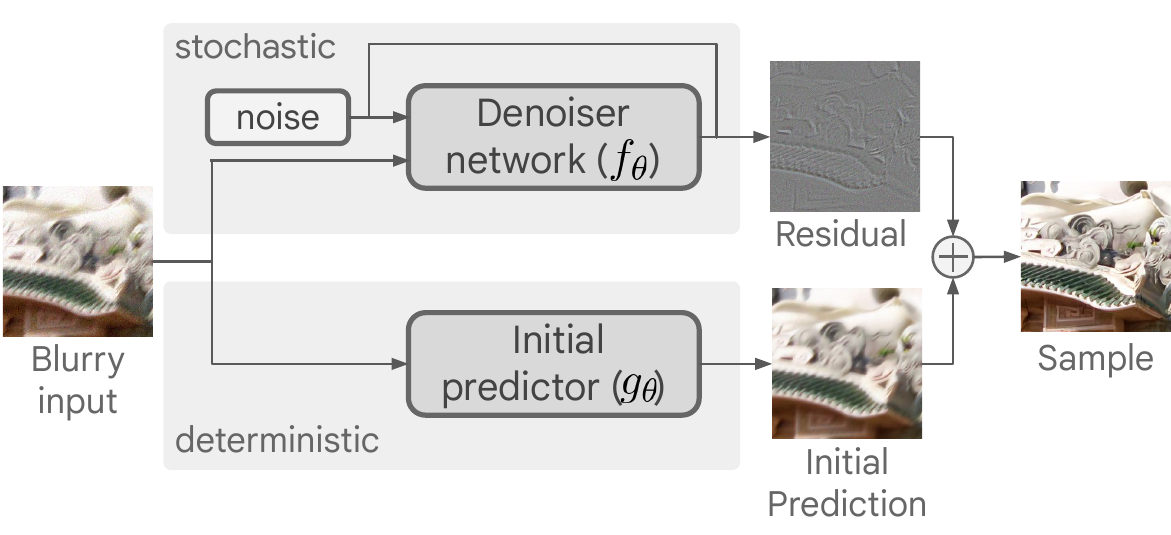}
    \vspace{-1em}
    \caption{Diagram describing our dual-network architecture. The initial predictor produces the deterministic candidate for the denoiser network, which then models the residual.}
    \label{fig:residual_diagram}
    \vspace{-1em}
\end{figure}

%% file: figures/fig_proposal_and_samples.tex
\begin{figure}[t]
    \centering
    \includegraphics[width=\linewidth]{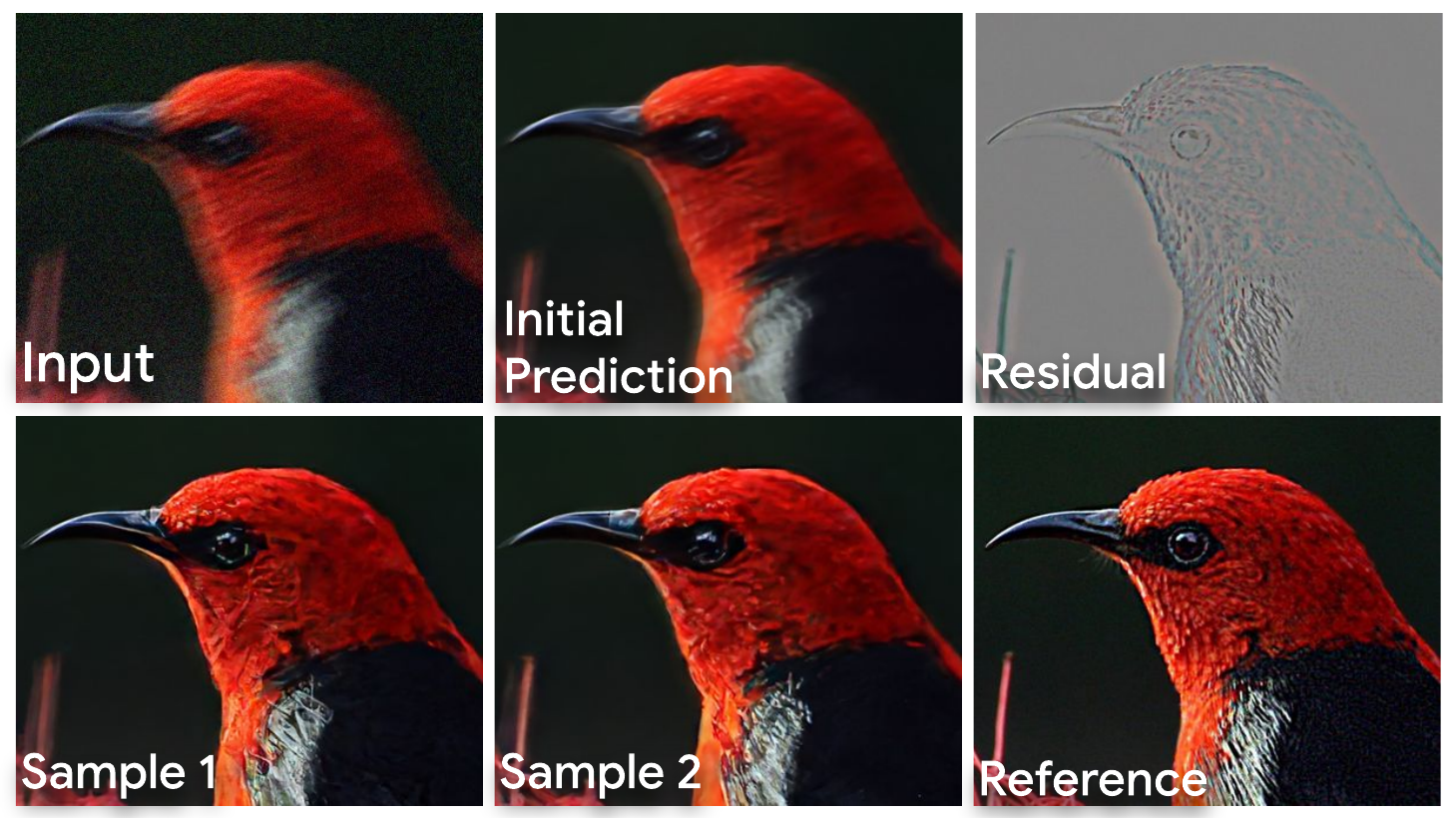}
    \vspace{-1.0em}
    \caption{Output of the initial predictor and multiple samples generated from it. We see that the over-smoothed initial prediction lacking texture is ``corrected'' by the stochastic sampler, producing crisp and diverse final reconstructions. The residual (top right) shows the the difference between reference and initial prediction.}
    \label{fig:proposal_and_samples}
\end{figure}

%% file: algorithms/residual_sampling.tex
\begin{algorithm}[t]
\caption{Predict-and-refine diffusion sampling.\\
The expressions for $\bmu_t, \baralpha_t, \beta_t$ can be found in \cref{sec:background}.}
\label{alg:residual_sampling}
\begin{algorithmic}[1]
    \REQUIRE $\ftheta$: Denoiser network, $\gtheta$: Initial predictor, \\$\by$: Blurry input image, $\balpha_{1:T}$: Noise schedule.
    \STATE $\bx_{\text{init}} \gets \gtheta(\by)$ \hfill $\triangleright$ Initial prediction
    \STATE $\bz_T \sim \N(\bzero, \I)$ \hfill $\triangleright$ Run diffusion sampling
    \FOR{$t = T, \ldots, 1$}
        \STATE $\eps_t \sim \N(\bzero, \I)$
        
        \STATE $\bz_{t-1} \gets \bmu_t(\bz_t, \ftheta(\bz_t,\baralpha_t,\by)) + \beta_t\eps_t$ \\
        \hfill $\triangleright$ Reverse diffusion step; see \cref{eqn:forward_posterior}
    \ENDFOR
    \RETURN $\bx_{\text{init}} + \bz_0$
    \hfill $\triangleright$ Return the final restoration
\end{algorithmic}
\end{algorithm}

%% file: tables/table_gopro_results_concise.tex
\begin{table}[ht]
\small
\setlength{\tabcolsep}{1pt}
\centering
\caption{Image deblurring results on the GoPro \cite{nah2017deep} dataset.  Our proposed method sets the new Pareto frontier in terms of Perception-Distortion trade-off. \colorbox{blue!15}{Best values} and \colorbox{green!15}{second-best values} for each metric are color-coded. KID values are scaled by a factor of 1000 for readability.}
\vspace{-0.5em}
\label{table:gopro_results}
\begin{tabular}{lcccccc}
\toprule

& \multicolumn{4}{c}{\textbf{Perceptual}}
& \multicolumn{2}{c}{\textbf{Distortion}} \\

\cmidrule(lr){2-5} \cmidrule(lr){6-7}

& LPIPS$\downarrow$
& NIQE$\downarrow$
& FID$\downarrow$
& KID$\downarrow$
& PSNR$\uparrow$
& SSIM$\uparrow$ \\

\midrule

Ground Truth
& 0.0   & 3.21         & 0.0       & 0.0
& $\infty$  & 1.000 \\

\cmidrule{1-7}

HINet \cite{chen2021hinet}
& 0.088    & 4.01     & 17.91     & 8.15
& \colorbox{green!15}{32.77}    & \colorbox{green!15}{0.960 } \\

MPRNet \cite{zamir2021multi}
& 0.089    & 4.09     & 20.18     & 9.10
& 32.66     & 0.959  \\

MIMO-UNet+ \cite{cho2021rethinking}
& 0.091    & 4.03     & 18.05     & 8.17
& 32.45     & 0.957  \\

SAPHNet \cite{suin2020spatially}
& 0.101    & \colorbox{green!15}{3.99}     & 19.06     & 8.48
& 31.89    & 0.953 \\

SimpleNet \cite{li2021perceptual}
& 0.108    &           &           &    
& 31.52     & 0.950 \\

DeblurGANv2 \cite{kupyn2019deblurgan}
& 0.117    & 3.68     & \colorbox{green!15}{13.40}     & \colorbox{green!15}{4.41}
& 29.08    & 0.918 \\

\cmidrule{1-7}

Ours
& \colorbox{blue!15}{0.059}  & \colorbox{blue!15}{3.39}  
& \colorbox{blue!15}{4.04} & \colorbox{blue!15}{0.98}
& 31.66    & 0.948 \\

Ours-SA
& \colorbox{green!15}{0.078}   & 4.07
& 17.46    & 8.03
& \colorbox{blue!15}{33.23}   & \colorbox{blue!15}{0.963} \\

\bottomrule

\end{tabular}
\end{table}

%% file: figures/fig_gopro_and_hide_samples_arxiv.tex
\begin{figure*}[ht]
    \centering
    \includegraphics[width=0.9\linewidth]{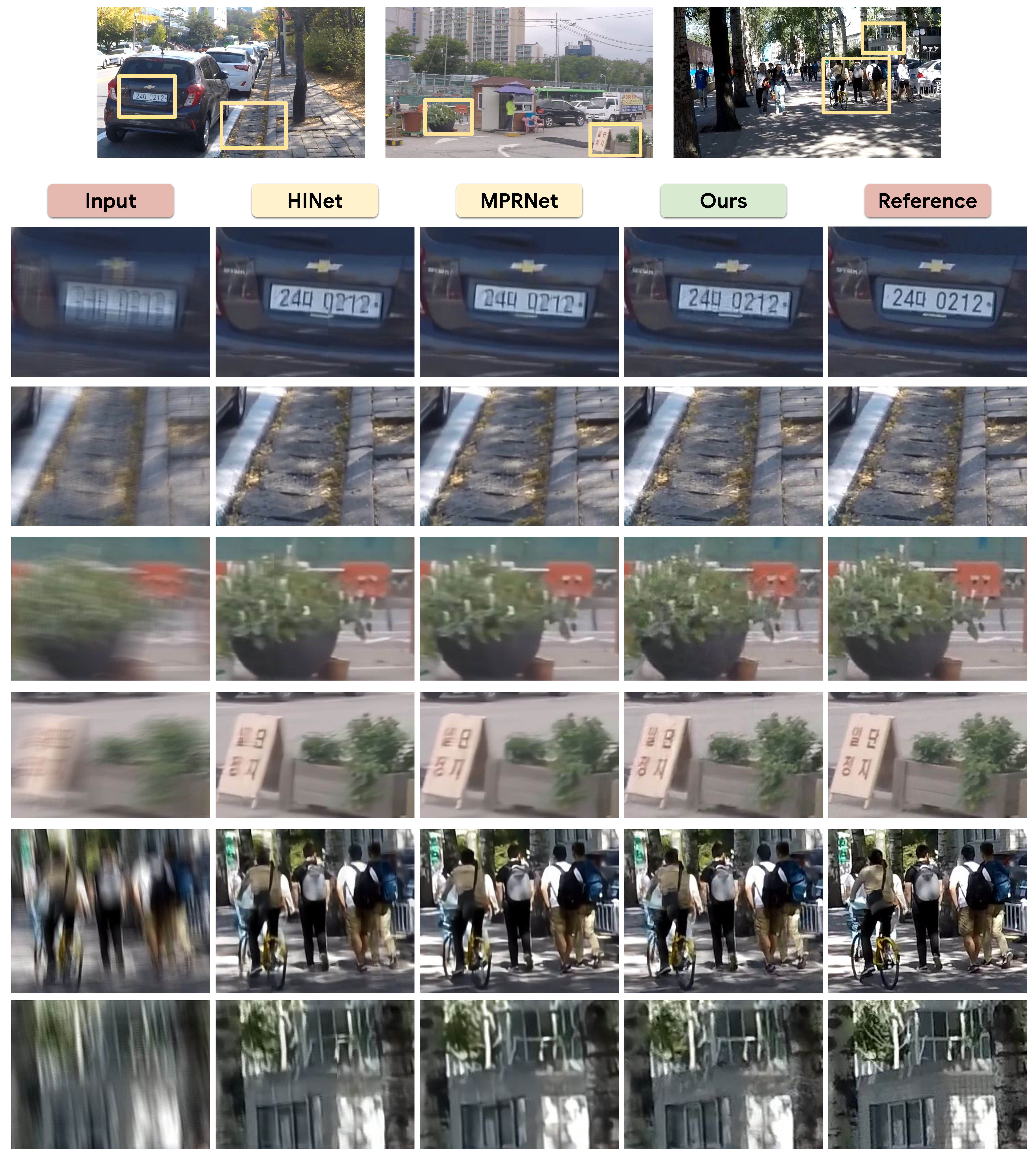}
    \vspace{-1em}
    \caption{Sample deblurred images from GoPro and HIDE datasets. Because our method is not trained to minimize distortion-based loss (\eg $L_2$), it avoids producing blurry output and achieves better reconstruction of detailed textures. Full-size images are provided in \Cref{supp:large_results}. %
    Best viewed electronically.
    }
    \label{fig:gopro_and_hide_samples}
\end{figure*}

%% file: tables/table_hide_results_concise.tex
\begin{table}[ht]
\small
\setlength{\tabcolsep}{1pt}

\centering
\caption{Image deblurring results on the HIDE \cite{shen2019human} dataset, using models trained on GoPro \cite{nah2017deep}.  Our method significantly outperforms the baseline methods under all perceptual metrics while maintaining competitive PSNR and SSIM.  \colorbox{blue!15}{Best values} and \colorbox{green!15}{second-best values} for each each metric are color-coded.}
\vspace{-0.5em}
\label{table:hide_results}
\begin{tabular}{lcccccc}
\toprule

& \multicolumn{4}{c}{\textbf{Perceptual}}
& \multicolumn{2}{c}{\textbf{Distortion}} \\

\cmidrule(lr){2-5} \cmidrule(lr){6-7}

& LPIPS$\downarrow$
& NIQE$\downarrow$
& FID$\downarrow$
& KID$\downarrow$
& PSNR$\uparrow$
& SSIM$\uparrow$ \\

\midrule

Ground Truth
& 0.0   & 2.72        & 0.0       & 0.0 
& $\infty$  & 1.000 \\

\cmidrule{1-7}

HINet \cite{chen2021hinet}
& 0.120   & 3.20     & 15.17    & 7.33
& \colorbox{green!15}{30.33}    & \colorbox{green!15}{0.932} \\

MIMO-UNet+ \cite{cho2021rethinking}
& 0.124   & 3.24     & 16.01    & 7.91
& 29.99    & 0.930 \\

MPRNet \cite{zamir2021multi}
& 0.114   & 3.46     & 16.58    & 8.35
& \colorbox{blue!15}{30.96}   & \colorbox{blue!15}{0.939} \\

SAPHNet \cite{suin2020spatially}
& 0.128   & 3.21     & 16.77    & 8.39
& 29.99    & 0.930 \\

DeblurGAN-v2 \cite{kupyn2019deblurgan}
& 0.159   & 2.96     & 15.51    & 6.97
& 27.51    & 0.885 \\

\cmidrule{1-7}

Ours
& \colorbox{blue!15}{0.089}   & \colorbox{blue!15}{2.69}
& \colorbox{blue!15}{5.43}    & \colorbox{blue!15}{1.61}
& 29.77    & 0.922 \\

Ours-SA

& \colorbox{green!15}{0.092}    & \colorbox{green!15}{2.93}
& \colorbox{green!15}{6.37}     & \colorbox{green!15}{2.40}
& 30.07    & 0.928 \\

\bottomrule

\end{tabular}
\vspace{-2em}
\end{table}

%% file: tables/table_perceptual_study.tex
\begin{table}[ht]
\setlength{\tabcolsep}{2pt}
\renewcommand{\arraystretch}{0.65}
    \centering
    \caption{Average pairwise human preference for deblurring results on the GoPro dataset~\cite{nah2017deep}. Each value represents the percentage of times Amazon Mechanical Turk raters chose the row over the column. Each preference percentage is an average over 480 ratings (20 raters, and 24 unique image pairs).}
    \vspace{-0.5em}
    \label{table:perceptual_study}
    \begin{tabular}{lcccccc} \toprule
             & HINet  & MPRNet	& Ours & Ours-SA & Reference \\\midrule
     HINet~\cite{chen2021hinet}  & - & 54.9& 29.1 & 31.0 & 14.5 \\
     MPRNet \cite{zamir2021multi}  & 45.1 &	- & 26.6 & 25.3 & 11.9	 \\
     Ours & \colorbox{blue!15}{70.9} &	\colorbox{blue!15}{73.4} &	- & 58.8 & 37.1 \\
     Ours-SA & \colorbox{blue!15}{69.0} & \colorbox{blue!15}{74.7} & 41.2 & - & 26.7 \\
     
     \cmidrule{1-6}
     
     Reference & 85.5 & 88.1 & 62.9 & 73.3 & -
     \\\midrule
    \end{tabular}
\end{table}

%% file: figures/fig_diversity_std.tex
\begin{figure}[t]
    \centering
    \includegraphics[width=\linewidth]{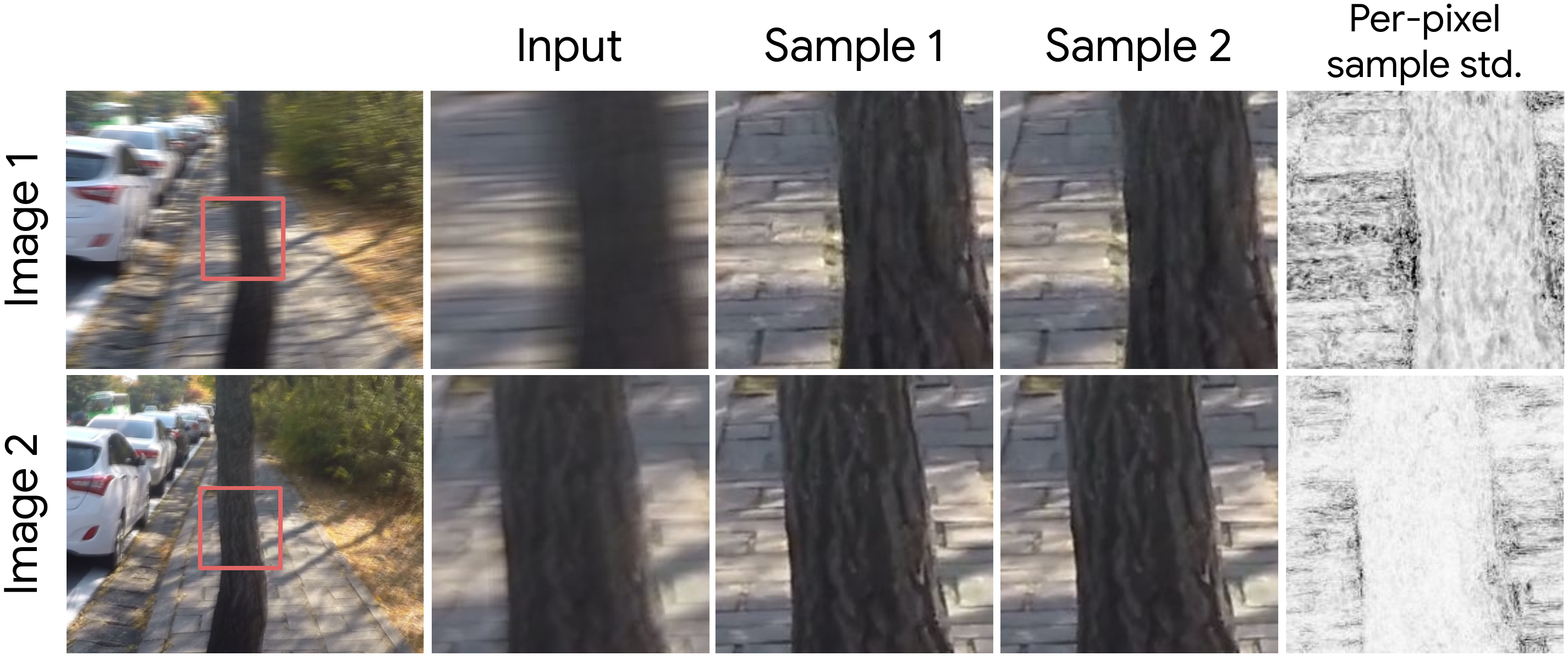}
    \vspace{-0.5em}
    \caption{Deblurred samples for crops of two different images. The ill-posedness of the restoration task (\ie strength of the blur) has a direct impact on the diversity of the generated samples. This is illustrated by the per-pixel standard deviation computed using multiple restorations for each input image. As clearly visible in the right-most column, the blurrier input (first row) corresponds to overall higher per-pixel standard deviations.}
    \label{fig:diversity_std}
    \vspace{-1em}
\end{figure}

%% file: figures/fig_sampling_cost.tex
\begin{figure}[t]
    \centering
    \includegraphics[width=\linewidth]{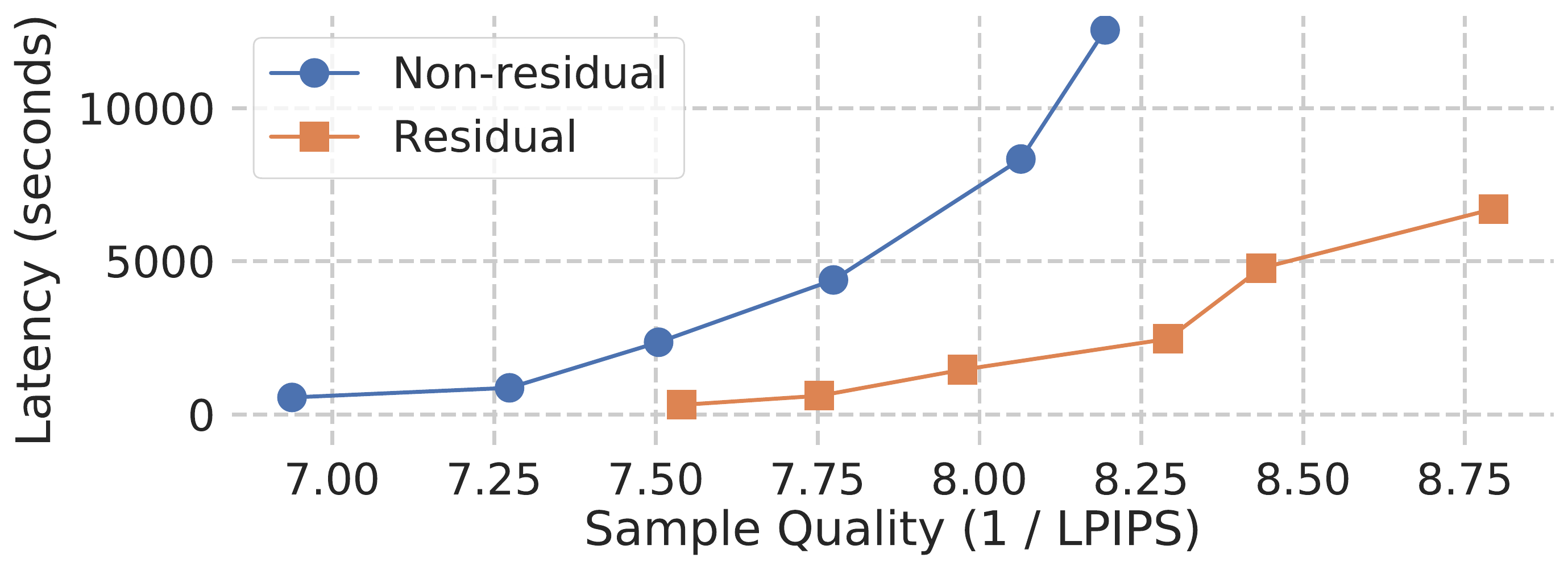}
    \vspace{-0.5em}
    \caption{Plot of sampling cost vs. sample quality. Even with the added parameters from the initial predictor, the residual model achieves lower latency while maintaining higher sample quality.
    }
    \label{fig:sampling_cost}
    \vspace{-0.5em}
\end{figure}

%% file: tables/table_regressor_ablation_concise.tex
\begin{table}[ht]
\small
\setlength{\tabcolsep}{2pt}
\centering
\caption{Ablation study on the effects of various hyperparameters for our U-Net architecture, evaluated on the GoPro dataset.
}
\vspace{-0.5em}
\label{table:regressor_ablation}
\begin{tabular}{lccccccc}
\toprule

& \multicolumn{3}{c}{Hyperparameters}  & \multicolumn{4}{c}{Metrics} \\

\cmidrule(lr){2-4} \cmidrule(lr){5-8}

& ch. & batch & EMA
& LPIPS & PSNR & MParam. & BFLOPs \\

\midrule

\multirow{3}{4em}{More Channels}
    & 16  & 32  & No & 0.137 & 29.93 & 1.63  & 301  \\
    & 32  & 32  & No & 0.113 & 31.05 & 6.52  & 1200 \\
    & 64  & 32  & No & 0.103 & 31.63 & 26.07 & 4790 \\
    
\midrule
    
\multirow{3}{4em}{$+$Larger Batch}
    & 64  & 64  & No & 0.099 & 31.85 & 26.07 & 4790 \\
    & 64  & 128 & No & 0.087 & 32.56 & 26.07 & 4790 \\
    & 64  & 256 & No & 0.086 & 32.61 & 26.07 & 4790 \\
    
\midrule

$+$Use EMA & 64  & 256 & Yes & 0.0809 & 33.07 & 26.07 & 4790 \\

\bottomrule

\end{tabular}
\end{table}

%% file: figures/sm/fig_additional_pd_plots.tex
\begin{figure*}[ht]
    \centering
    \includegraphics[width=0.9\linewidth]{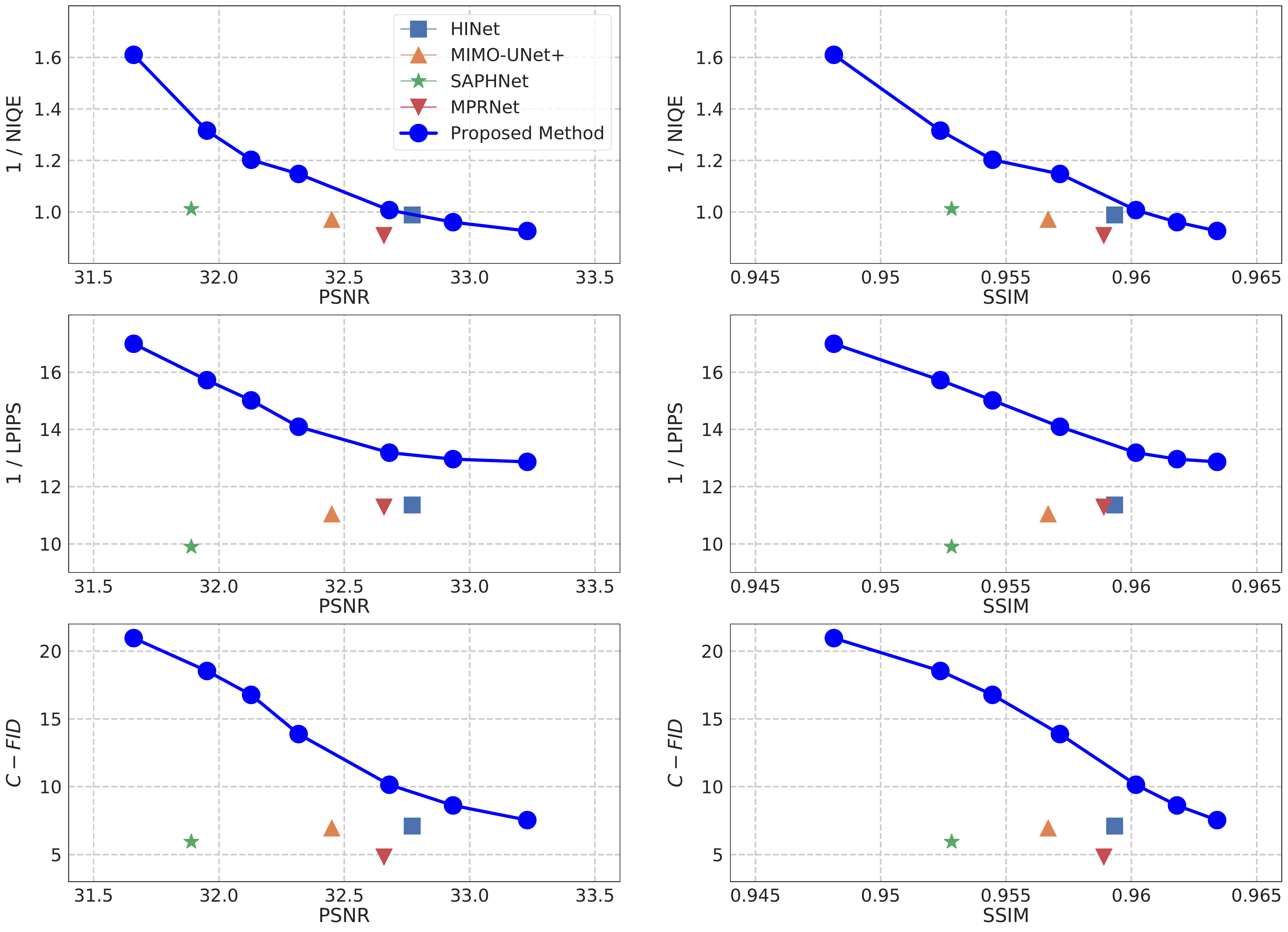}
    \caption{Additional Perception-Distortion plots with respect to different metrics.  Left column contains perceptual metrics vs. PSNR, and the right column contains SSIM comparisons. We notice that the same trade-off is present for all (perceptual, distortion) metric pairs.
    }
    \label{fig:additional_pd_plots}
\end{figure*}

%% file: figures/sm/fig_diversity_std_blur_gopro.tex
\begin{figure*}[ht]
    \begin{minipage}[c]{.29\linewidth}
    \includegraphics[width=\linewidth]{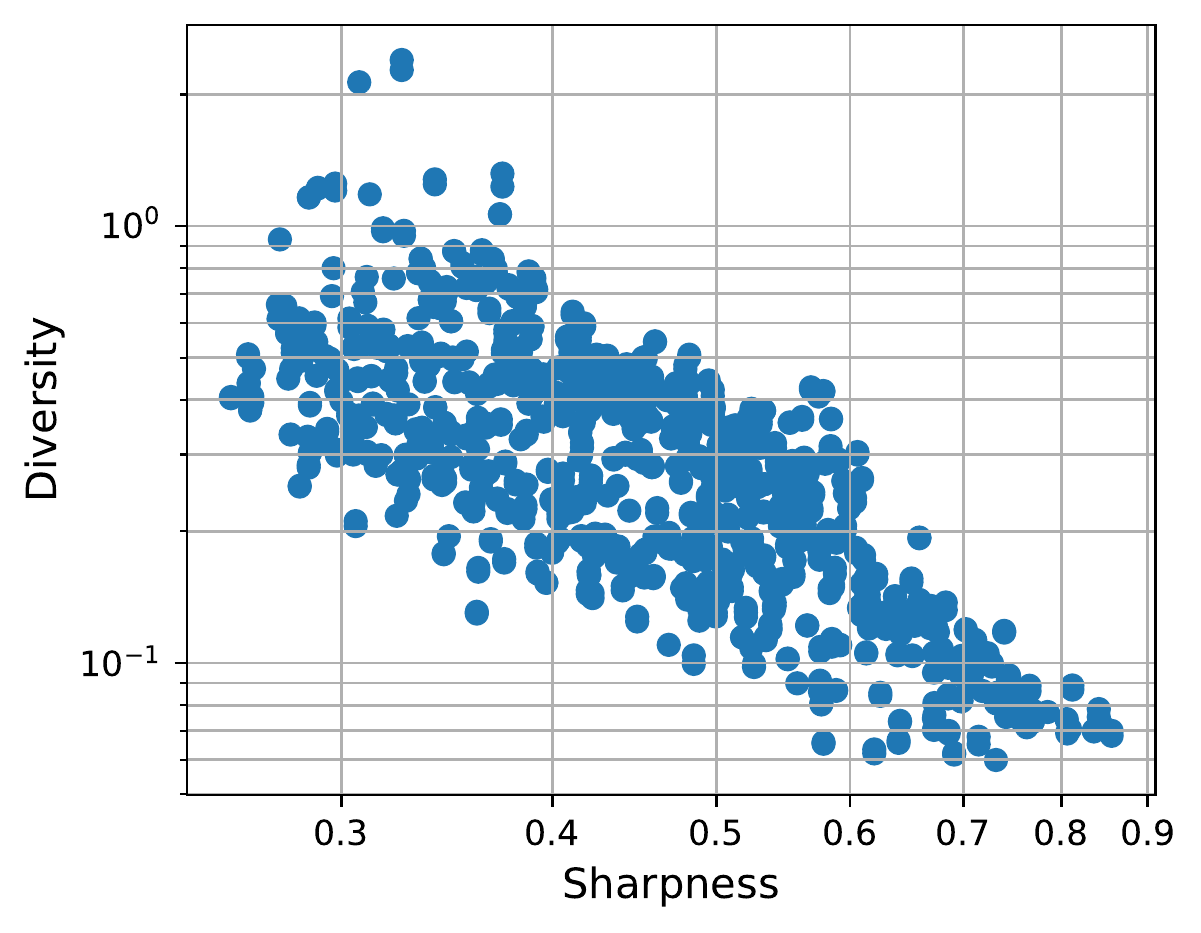}
    \end{minipage}
    \begin{minipage}[c]{.69\linewidth}
    \includegraphics[width=\linewidth]{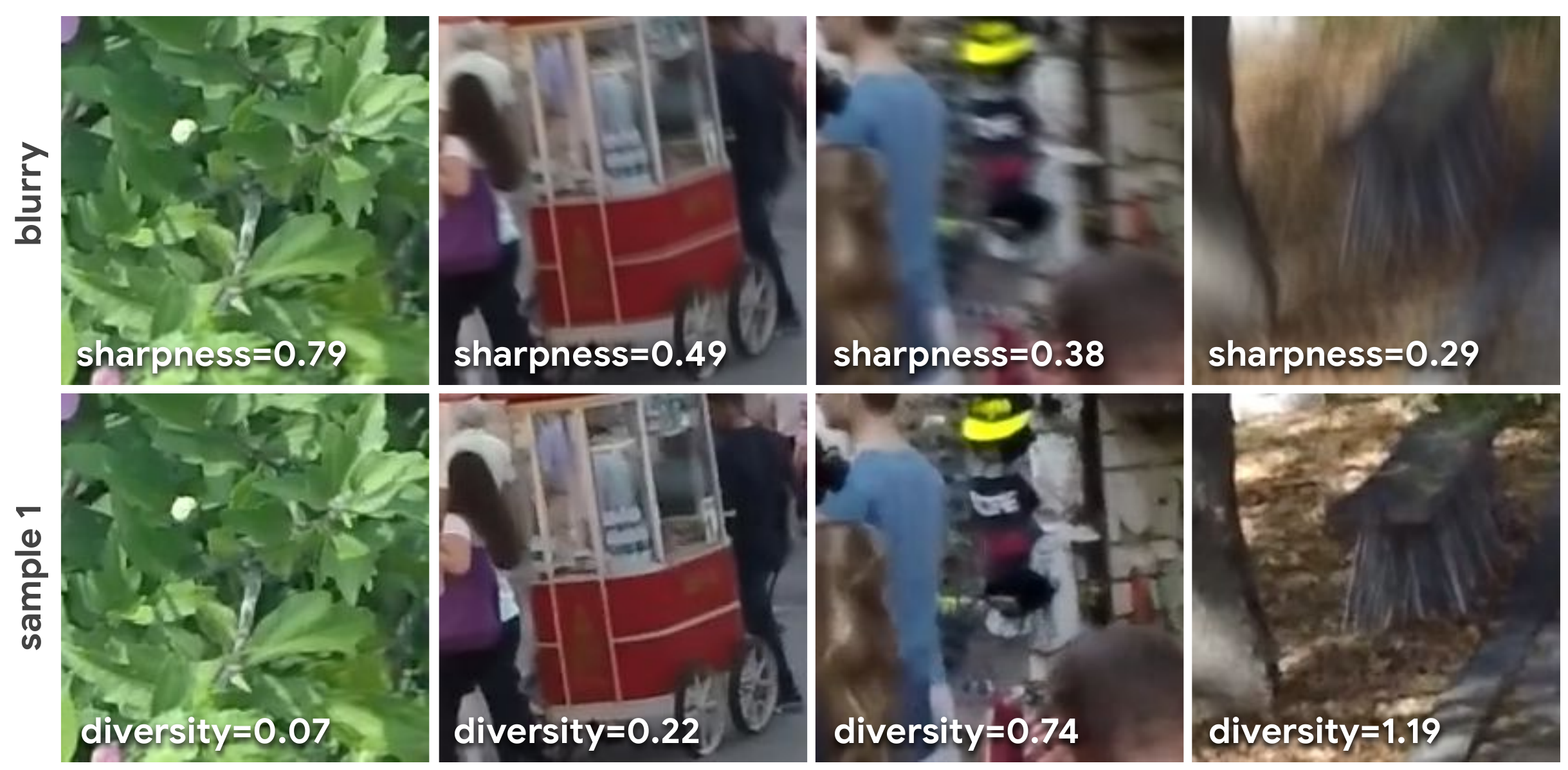}
    \end{minipage}
    \caption{Sample diversity as a function of input image sharpness. The ill-posedness of the restoration task (\ie how strong the blur is) has a direct impact on the diversity of the generated samples. \textbf{Left:} Each point in this plot represents an image in the GoPro validation set. Image sharpness is computed as: $\text{sharpness}=\|\Delta \text{input}\| / \|\Delta \text{reference}\|$, where $\Delta$ is the Laplacian of the given image. Sample diversity is computed as: $\text{diversity}=\|\text{Var}\left[\text{sample}\right]\|/\|\Delta \text{reference}\|$, where $\text{Var}[\text{sample}]$ is the per pixel empirical variance of multiple restored images for a given input. \textbf{Right:} Four different blurry image crops with different level of sharpness, and a respective deblurred sample for each one (sample 1).}
    \label{fig:diversity_std_blur_gopro}
\end{figure*}

%% file: figures/sm/fig_architecture_diagram.tex
\begin{figure*}[ht]
    \centering
    \includegraphics[width=.9\linewidth]{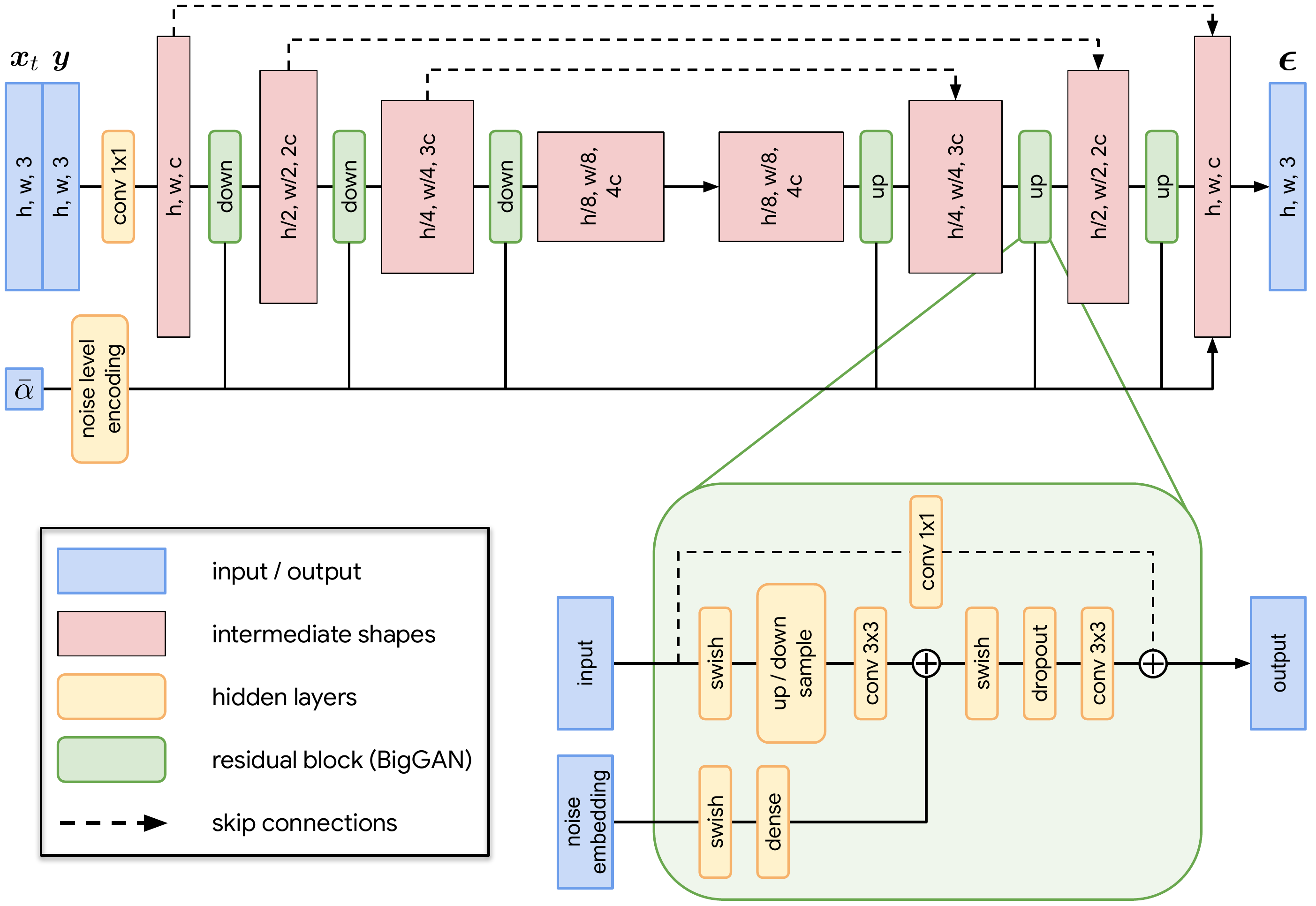}
    \caption{Diagram describing the U-Net architecture used for both the denoiser network and the initial predictor in our experiments. Note that the input and output depicted here are for the denoiser network.}
    \label{fig:architecture_diagram}
\end{figure*}

%% file: figures/sm/fig_gopro_fullsize_1.tex
\begin{figure*}[ht]
    \centering
    \includegraphics[width=0.99\linewidth]{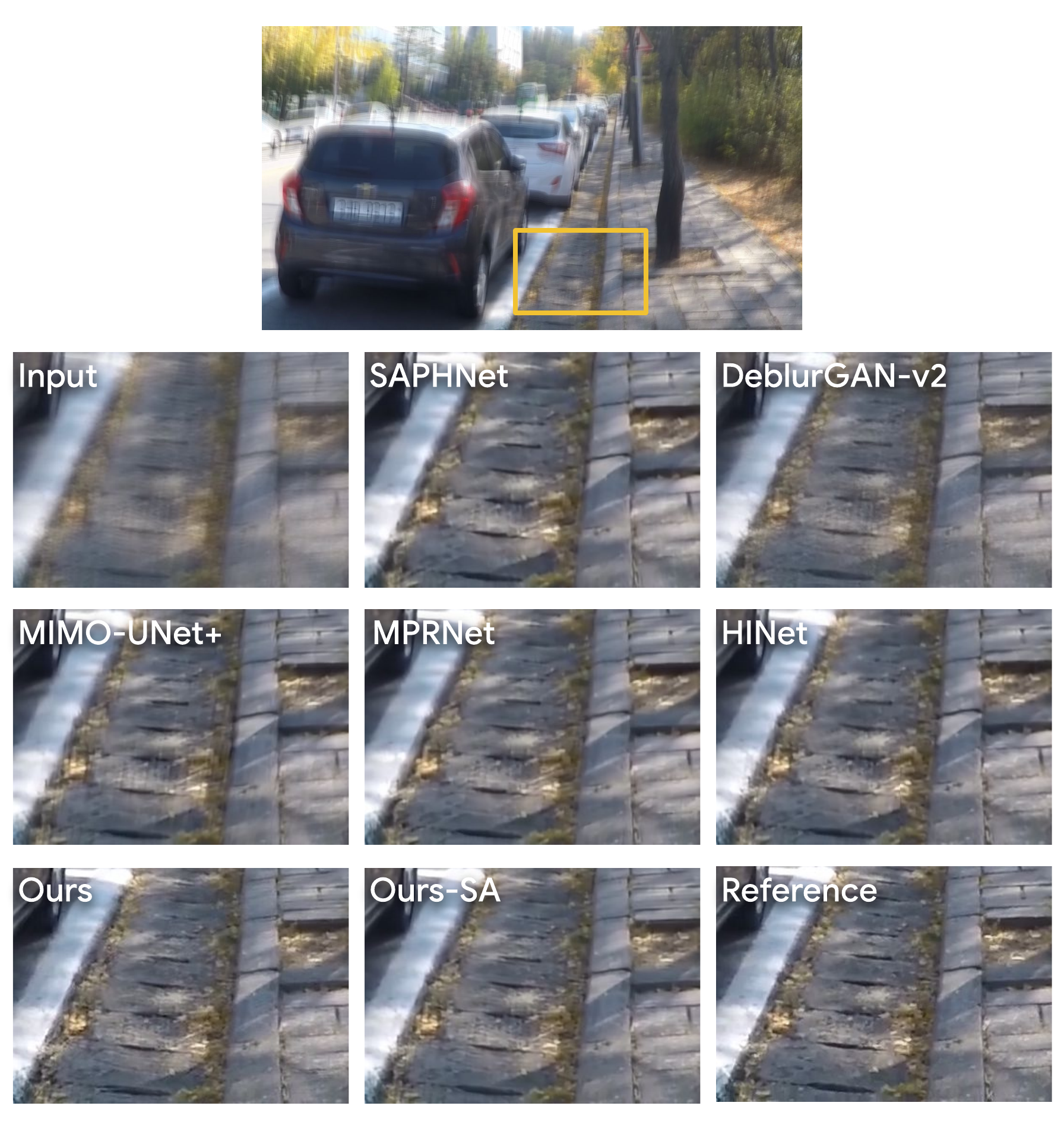}
    \caption{Full comparison of the GoPro \cite{nah2017deep} deblurring result presented in the main text. The compared methods are: SAPHNet~\cite{suin2020spatially}, DeblurGAN-v2~\cite{kupyn2019deblurgan}, MIMO-Unet+~\cite{cho2021rethinking}, MPRNet~\cite{zamir2021multi}, and HINet~\cite{chen2021hinet}.  We include restorations from our method with and without sampling averaging (``Ours'' and ``Ours-SA'').}
    \label{fig:supp_gopro_fullsize_1}
\end{figure*}

%% file: figures/sm/fig_gopro_fullsize_2.tex
\begin{figure*}[ht]
    \centering
    \includegraphics[width=0.99\linewidth]{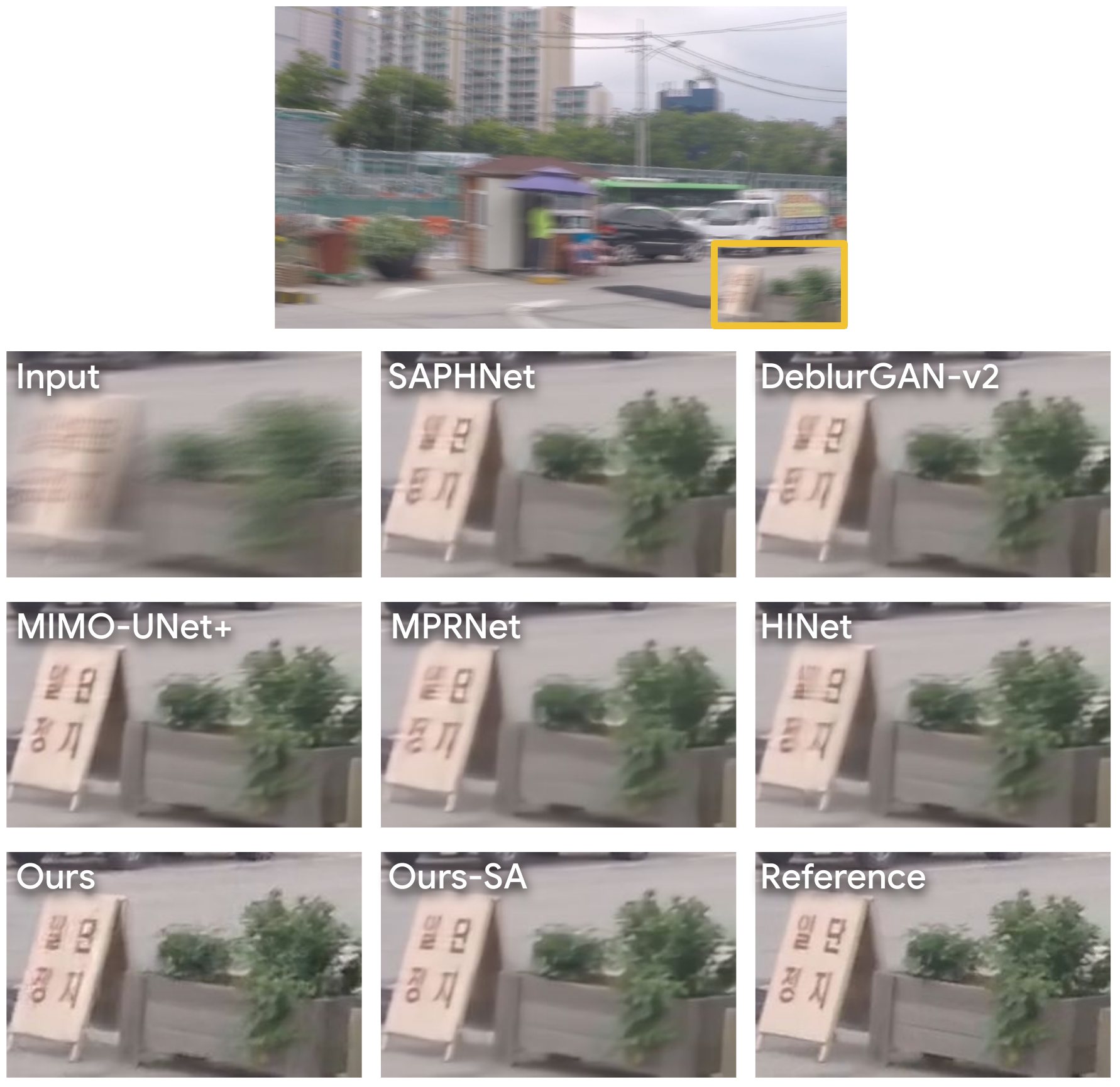}
    \caption{Full comparison of the GoPro \cite{nah2017deep} deblurring result presented in the main text. The compared methods are: SAPHNet~\cite{suin2020spatially}, DeblurGAN-v2~\cite{kupyn2019deblurgan}, MIMO-Unet+~\cite{cho2021rethinking}, MPRNet~\cite{zamir2021multi}, and HINet~\cite{chen2021hinet}.  We include restorations from our method with and without sampling averaging (``Ours'' and ``Ours-SA'').}
    \label{fig:supp_gopro_fullsize_2}
\end{figure*}

%% file: figures/sm/fig_hide_fullsize_1.tex
\begin{figure*}[ht]
    \centering
    \includegraphics[width=0.99\linewidth]{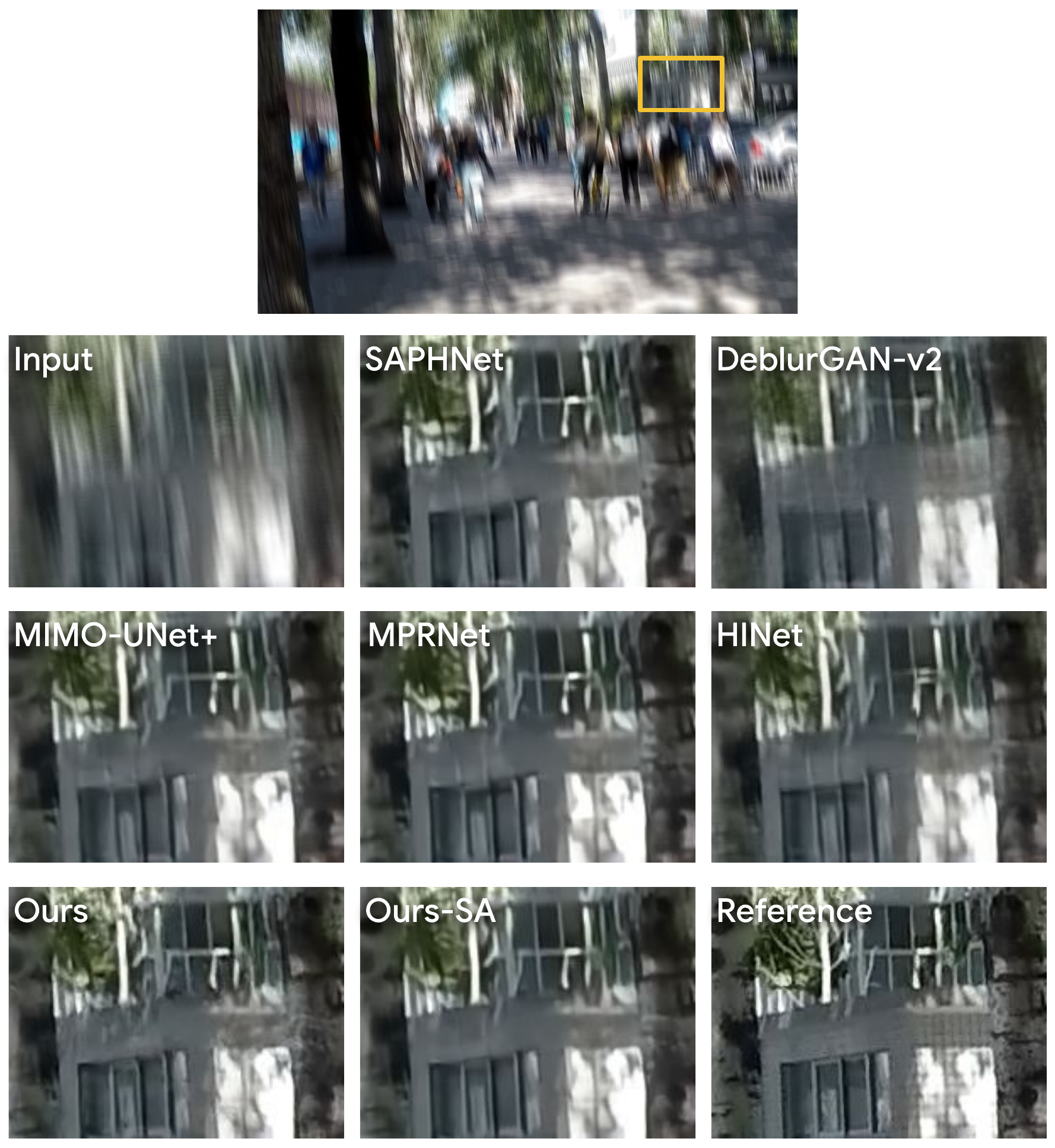}
    \caption{Full comparison of the HIDE \cite{shen2019human} deblurring result presented in the main text. The compared methods are: SAPHNet~\cite{suin2020spatially}, DeblurGAN-v2~\cite{kupyn2019deblurgan}, MIMO-Unet+~\cite{cho2021rethinking}, MPRNet~\cite{zamir2021multi}, and HINet~\cite{chen2021hinet}.  We include restorations from our method with and without sampling averaging (``Ours'' and ``Ours-SA'').}
    \label{fig:supp_hide_fullsize_1}
\end{figure*}

%% file: figures/sm/fig_gopro_samples_1.tex
\begin{figure*}[ht]
    \centering
    \includegraphics[width=0.9\linewidth]{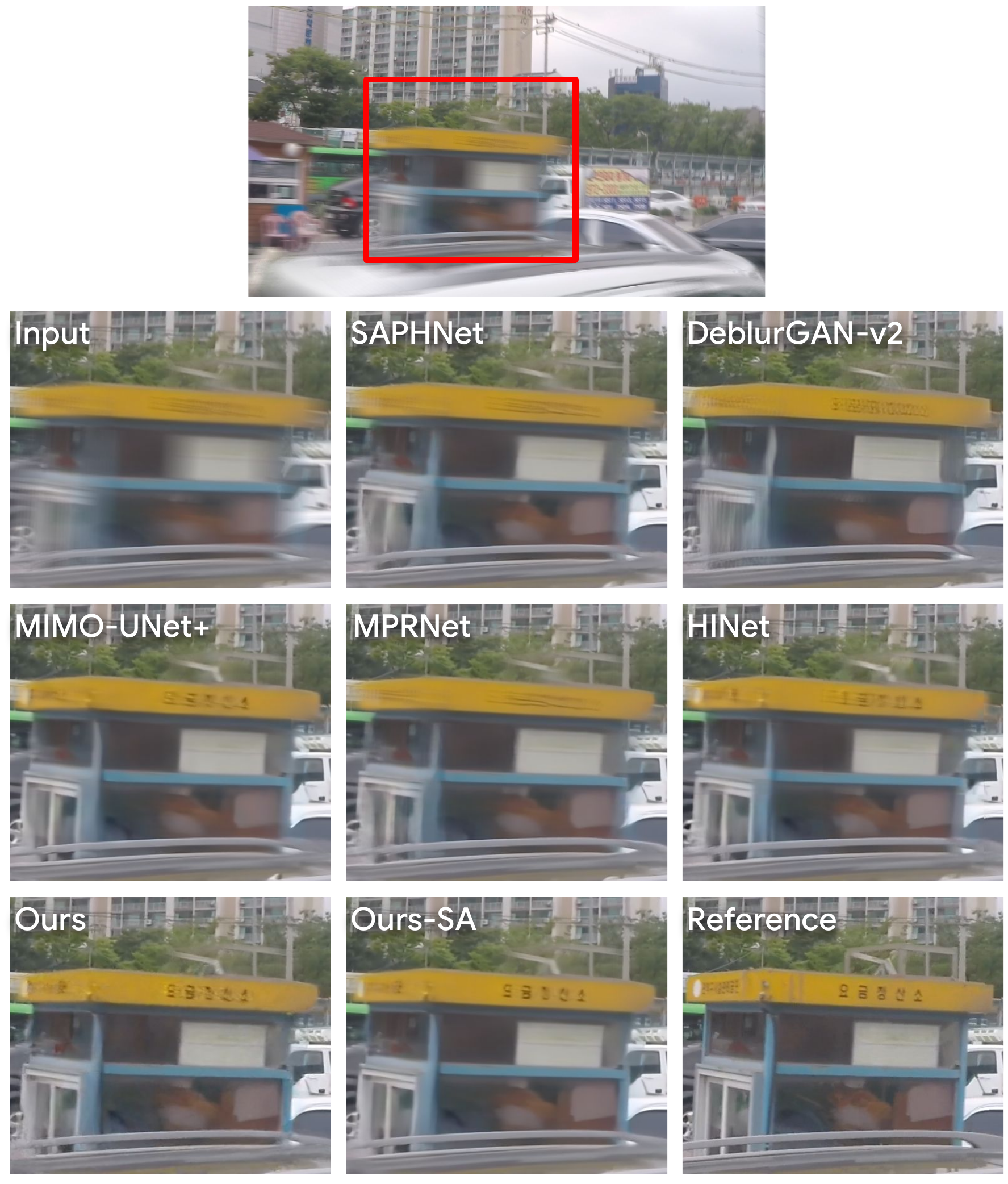}
    \caption{Additional deblurring results on the GoPro~\cite{nah2017deep} dataset. The compared methods are: SAPHNet~\cite{suin2020spatially}, DeblurGAN-v2~\cite{kupyn2019deblurgan}, MIMO-Unet+~\cite{cho2021rethinking}, MPRNet~\cite{zamir2021multi}, HINet~\cite{chen2021hinet}, and our method with and without sampling averaging.}
    \label{fig:supp_gopro_samples_1}
\end{figure*}

%% file: figures/sm/fig_gopro_samples_2.tex
\begin{figure*}[ht]
    \centering
    \includegraphics[width=\linewidth]{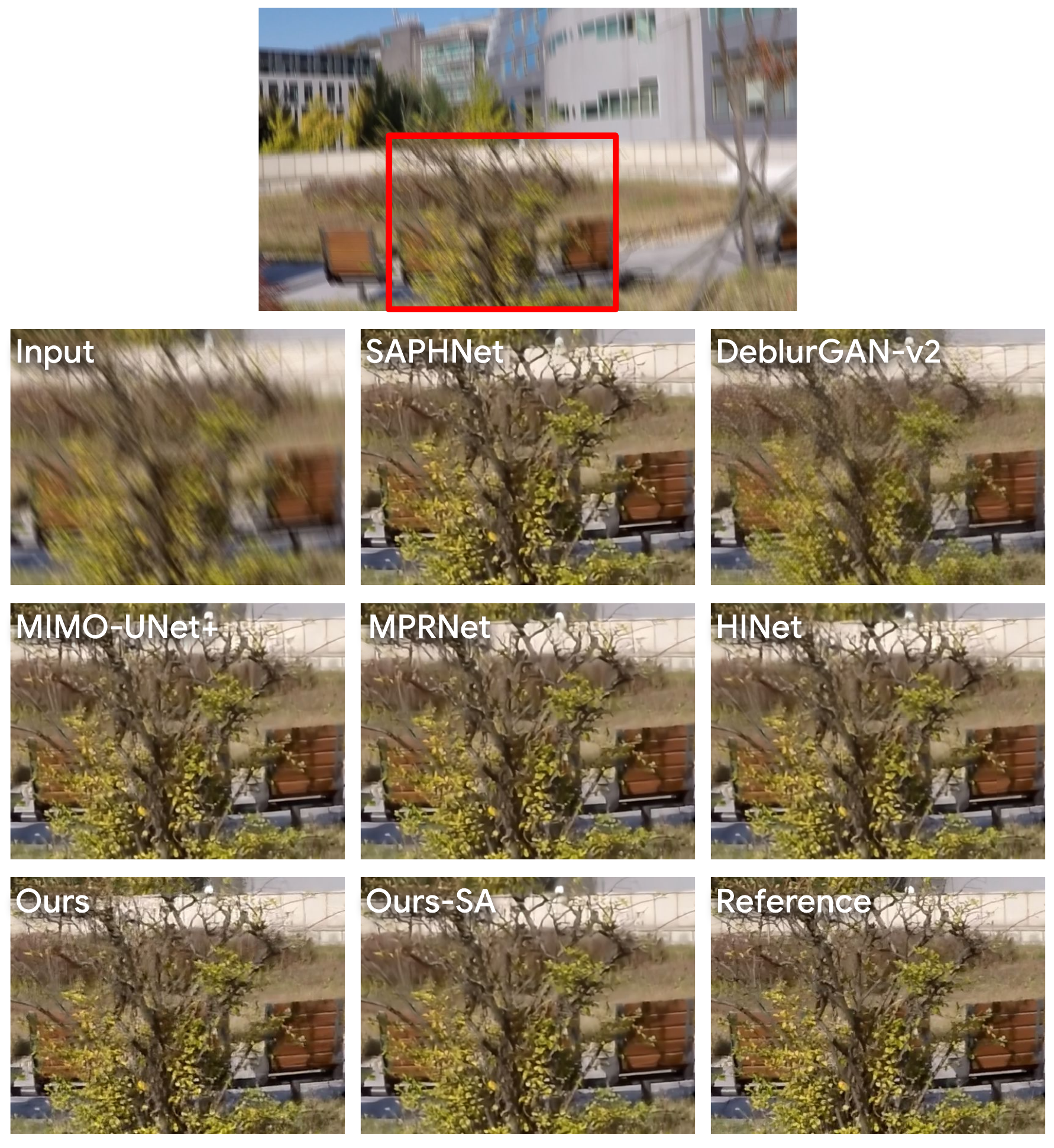}
    \caption{Additional deblurring results on the GoPro~\cite{nah2017deep} dataset. The compared methods are: SAPHNet~\cite{suin2020spatially}, DeblurGAN-v2~\cite{kupyn2019deblurgan}, MIMO-Unet+~\cite{cho2021rethinking}, MPRNet~\cite{zamir2021multi}, HINet~\cite{chen2021hinet}, and our method with and without sampling averaging.}
    \label{fig:supp_gopro_samples_2}
\end{figure*}

%% file: figures/sm/fig_gopro_samples_3.tex
\begin{figure*}[ht]
    \centering
    \includegraphics[width=\linewidth]{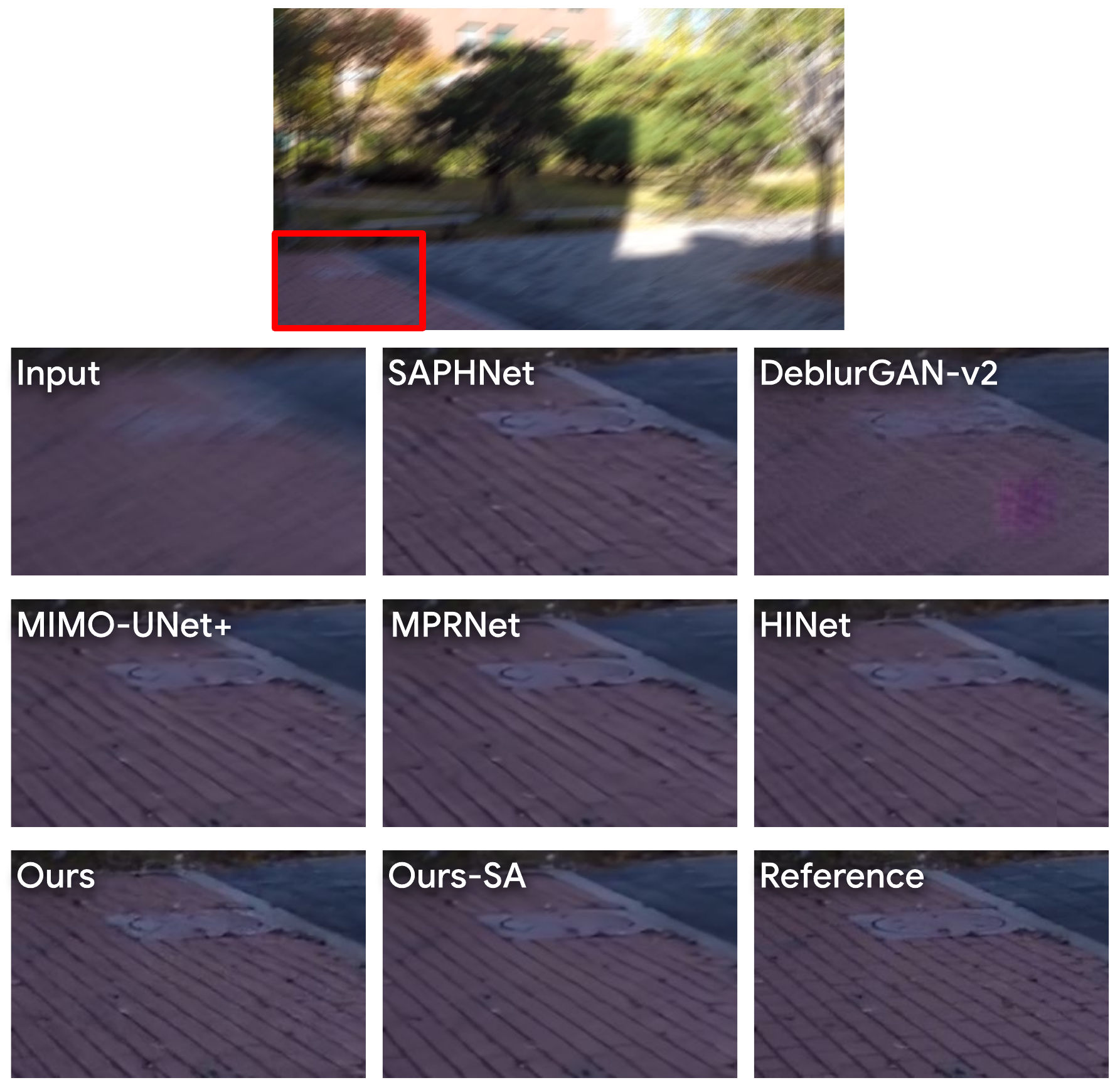}
    \caption{Additional deblurring results on the GoPro~\cite{nah2017deep} dataset. The compared methods are: SAPHNet~\cite{suin2020spatially}, DeblurGAN-v2~\cite{kupyn2019deblurgan}, MIMO-Unet+~\cite{cho2021rethinking}, MPRNet~\cite{zamir2021multi}, HINet~\cite{chen2021hinet}, and our method with and without sampling averaging.}
    \label{fig:supp_gopro_samples_3}
\end{figure*}

%% file: figures/sm/fig_gopro_samples_4.tex
\begin{figure*}[ht]
    \centering
    \includegraphics[width=\linewidth]{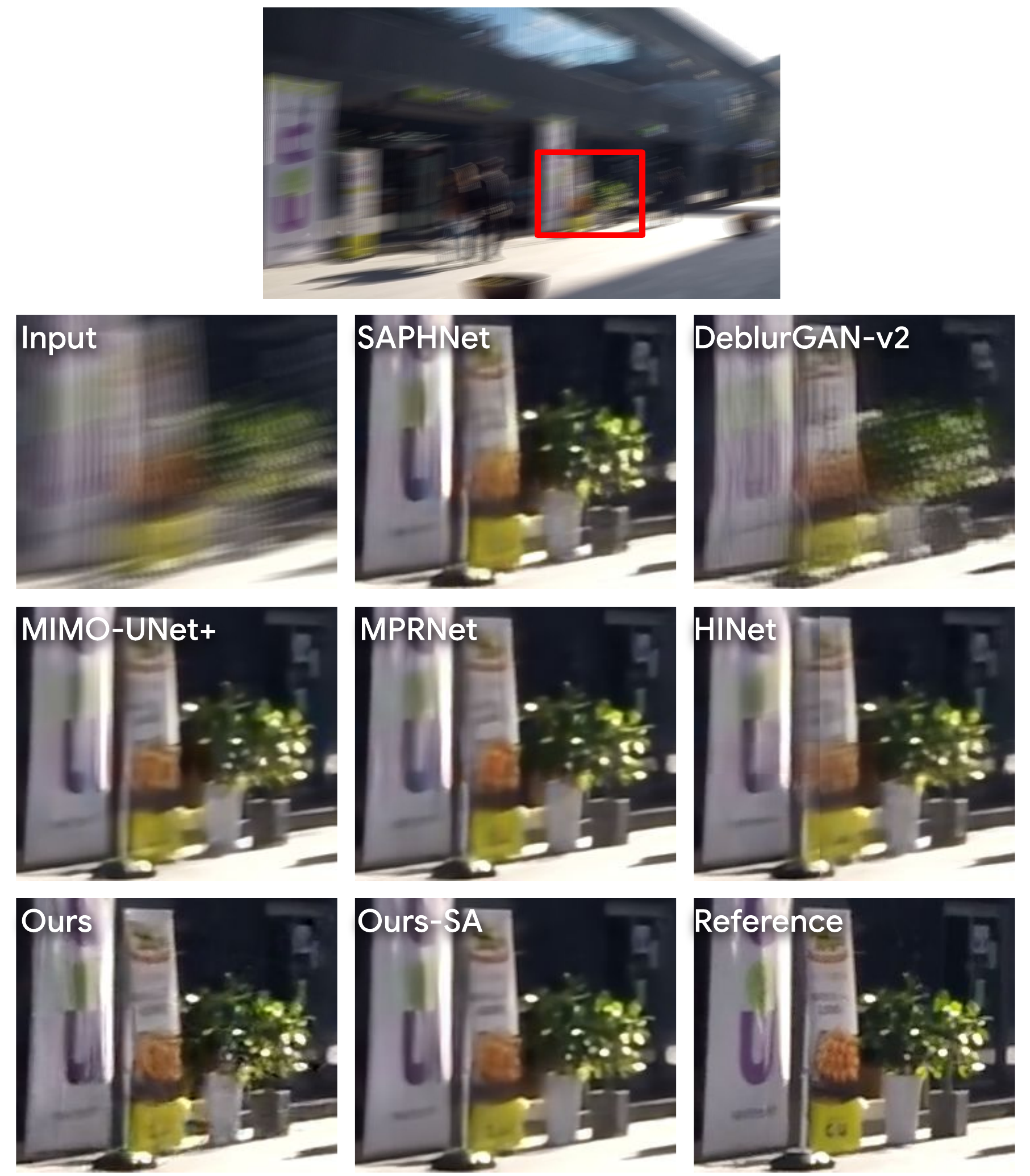}
    \caption{Additional deblurring results on the GoPro~\cite{nah2017deep} dataset. The compared methods are: SAPHNet~\cite{suin2020spatially}, DeblurGAN-v2~\cite{kupyn2019deblurgan}, MIMO-Unet+~\cite{cho2021rethinking}, MPRNet~\cite{zamir2021multi}, HINet~\cite{chen2021hinet}, and our method with and without sampling averaging.}
    \label{fig:supp_gopro_samples_4}
\end{figure*}

%% file: figures/sm/fig_gopro_samples_5.tex
\begin{figure*}[ht]
    \centering
    \includegraphics[width=\linewidth]{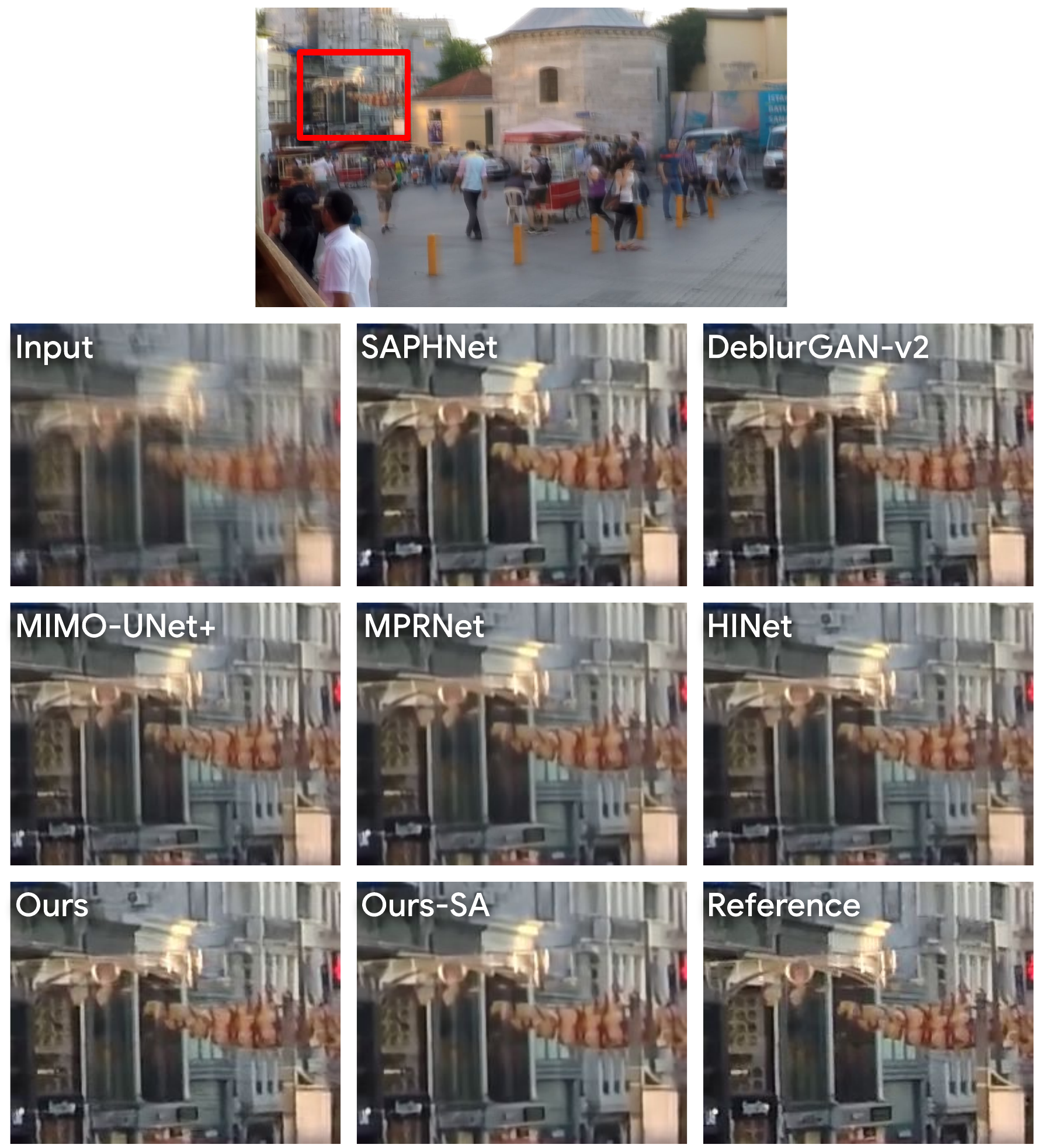}
    \caption{Additional deblurring results on the GoPro~\cite{nah2017deep} dataset. The compared methods are: SAPHNet~\cite{suin2020spatially}, DeblurGAN-v2~\cite{kupyn2019deblurgan}, MIMO-Unet+~\cite{cho2021rethinking}, MPRNet~\cite{zamir2021multi}, HINet~\cite{chen2021hinet}, and our method with and without sampling averaging.}
    \label{fig:supp_gopro_samples_5}
\end{figure*}

%% file: figures/sm/fig_div2k_samples_1.tex
\begin{figure*}[ht]
    \centering
    \includegraphics[width=0.9\linewidth]{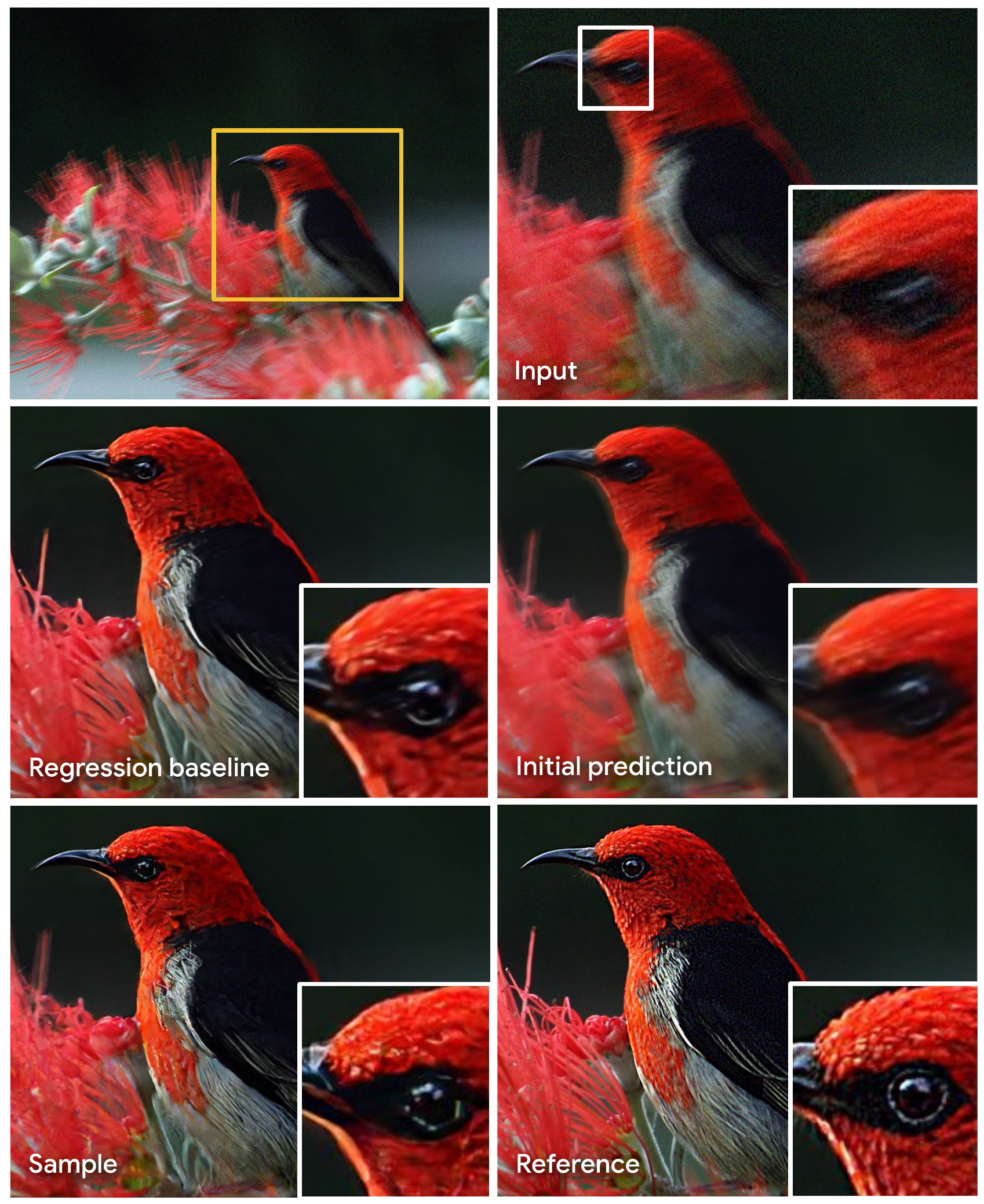}
    \caption{Additional deblurring results on the custom DIV2K dataset. We see that the initial predictor's blurry output is enhanced by the denoiser with realistic details.}
    \label{fig:supp_div2k_samples_1}
\end{figure*}

%% file: figures/sm/fig_div2k_samples_2.tex
\begin{figure*}[ht]
    \centering
    \includegraphics[width=1.0\linewidth]{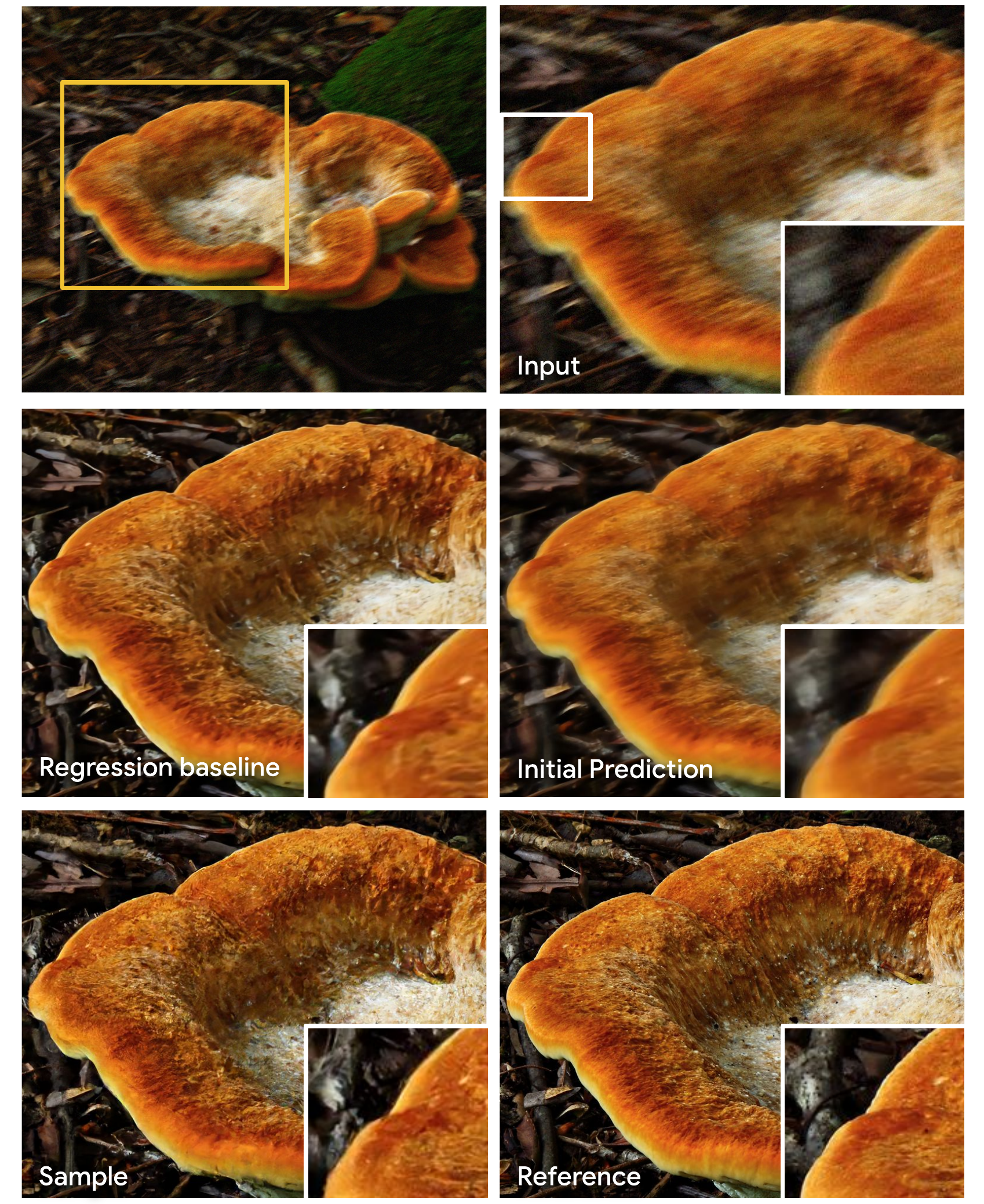}
    \caption{Additional deblurring results on the custom DIV2K dataset. We see that the initial predictor's blurry output is enhanced by the denoiser with realistic details.}
    \label{fig:supp_div2k_samples_2}
\end{figure*}

%% file: figures/sm/fig_div2k_samples_3.tex
\begin{figure*}[ht]
    \centering
    \includegraphics[width=0.9\linewidth]{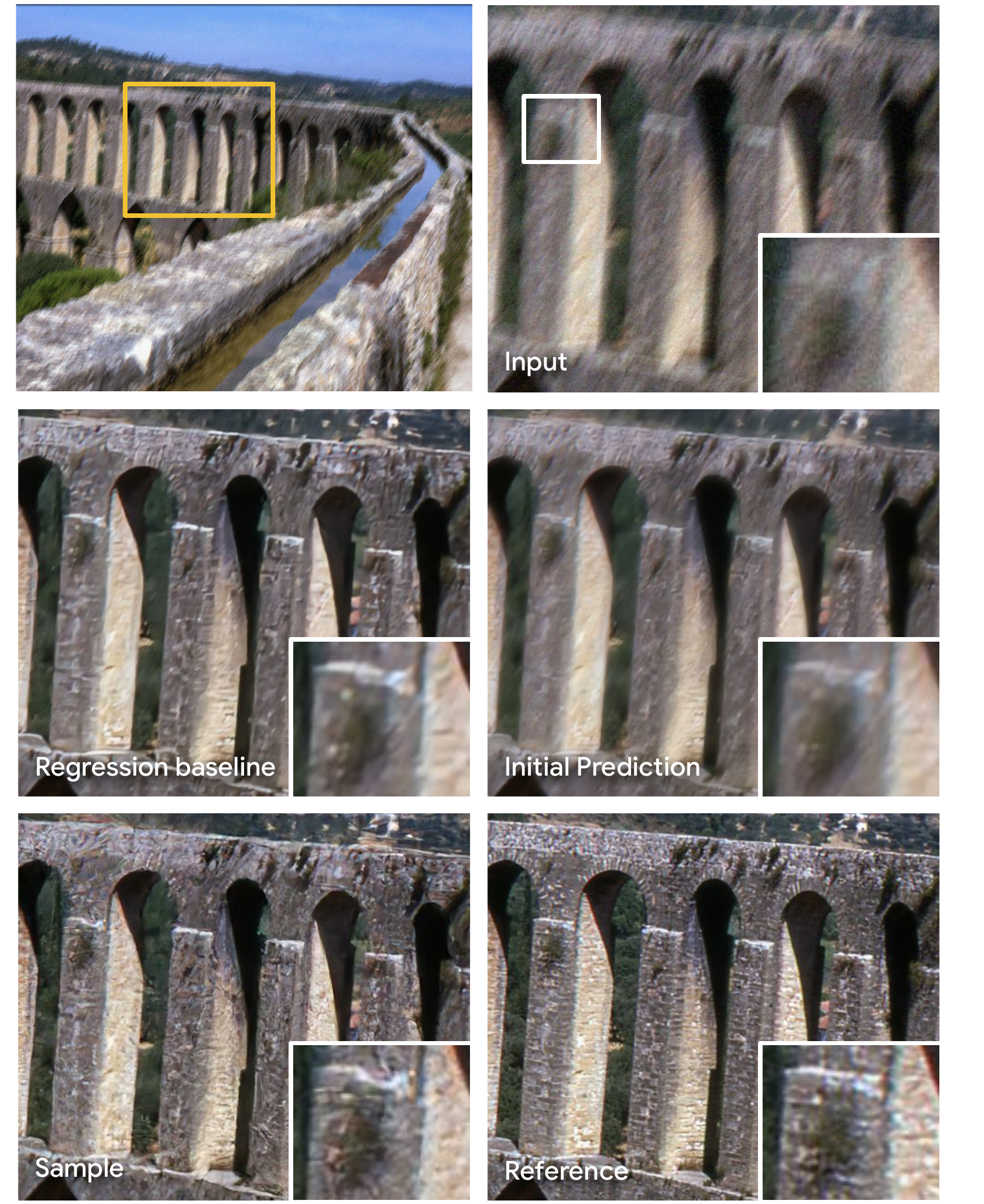}
    \caption{Additional deblurring results on the custom DIV2K dataset. We see that the initial predictor's blurry output is enhanced by the denoiser with realistic details.}
    \label{fig:supp_div2k_samples_3}
\end{figure*}

%% file: figures/sm/fig_div2k_samples_4.tex
\begin{figure*}[ht]
    \centering
    \includegraphics[width=0.9\linewidth]{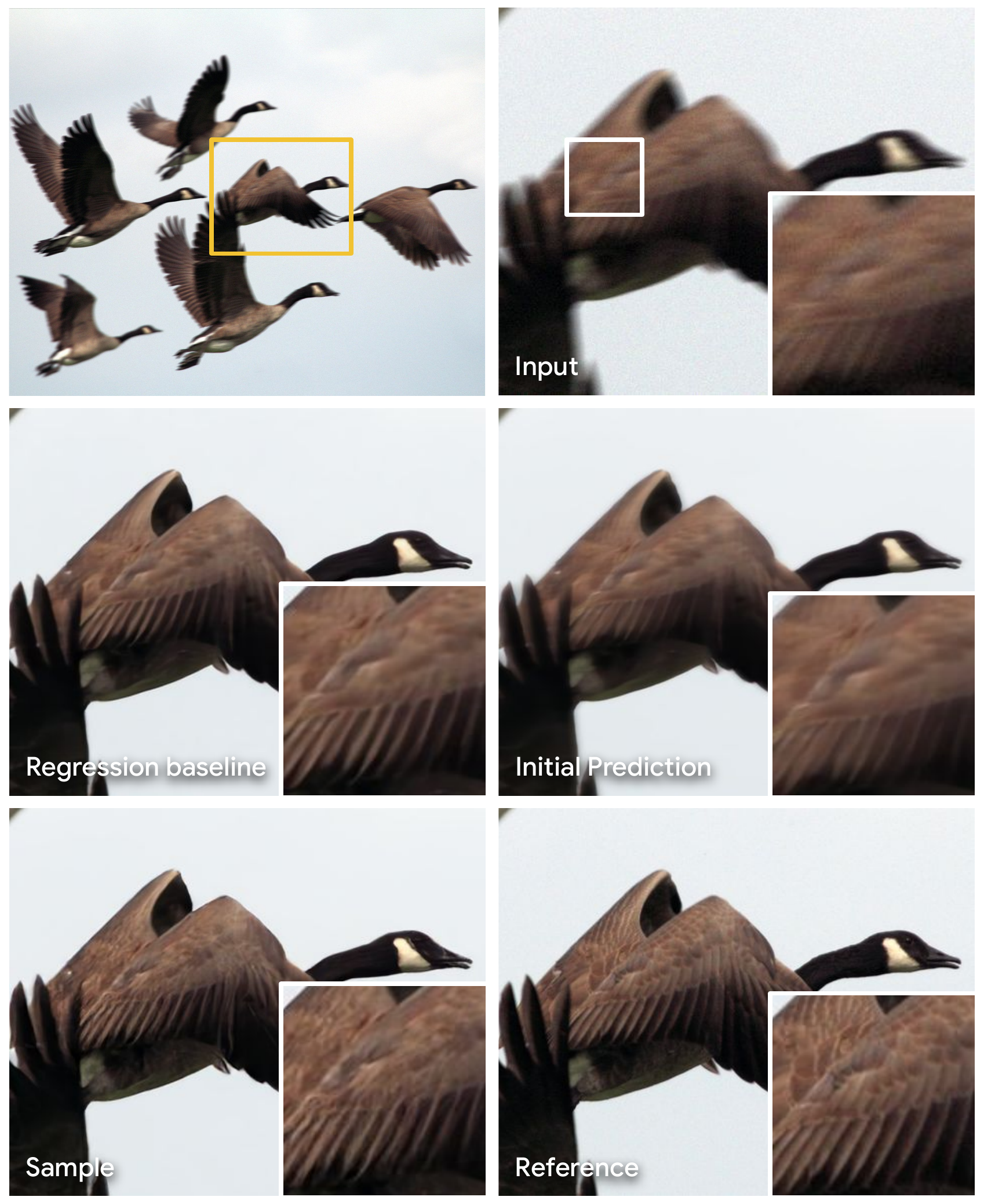}
    \caption{Additional deblurring results on the custom DIV2K dataset. We see that the initial predictor's blurry output is enhanced by the denoiser with realistic details.}
    \label{fig:supp_div2k_samples_4}
\end{figure*}